\documentclass[letterpaper]{article} % DO NOT CHANGE THIS
\usepackage{aaai2026}  % DO NOT CHANGE THIS
\usepackage{times}  % DO NOT CHANGE THIS
\usepackage{helvet}  % DO NOT CHANGE THIS
\usepackage{courier}  % DO NOT CHANGE THIS
\usepackage[hyphens]{url}  % DO NOT CHANGE THIS
\usepackage{graphicx} % DO NOT CHANGE THIS
\usepackage{natbib}  % DO NOT CHANGE THIS AND DO NOT ADD ANY OPTIONS TO IT
\usepackage{textcomp}
\usepackage{caption} % DO NOT CHANGE THIS AND DO NOT ADD ANY OPTIONS TO IT
\usepackage{array}
\usepackage{booktabs}
\usepackage{multirow}
\usepackage{multicol}
\usepackage{algorithm}
\usepackage{algorithmic}
\usepackage{longtable}
\usepackage{newfloat}
\usepackage{listings}
\DeclareCaptionStyle{ruled}{labelfont=normalfont,labelsep=colon,strut=off} % DO NOT CHANGE THIS
\floatstyle{ruled}
\newfloat{listing}{tb}{lst}{}
\floatname{listing}{Listing}
\title{Are LLMs Enough for Hyperpartisan, Fake, Polarized and Harmful Content Detection? Evaluating In-Context Learning vs. Fine-Tuning}

\title{Are LLMs Enough for Hyperpartisan, Fake, Polarized and Harmful Content Detection? Evaluating In-Context Learning vs. Fine-Tuning}
\author {
    Michele Joshua Maggini\textsuperscript{\rm 1},
    Dhia Merzougui\textsuperscript{\rm 2},
    Rabiraj Bandyopadhyay\textsuperscript{\rm 3},
    Gaël Dias\textsuperscript{\rm 2},
    Fabrice Maurel\textsuperscript{\rm 2},
    Pablo Gamallo\textsuperscript{\rm 1}
}
\affiliations {
    \textsuperscript{\rm 1}Centro Singular de Investigación en Tecnoloxías Intelixentes da USC, Universidade de Santiago de Compostela\\
    \textsuperscript{\rm 2}UNICAEN, ENSICAEN, CNRS, GREYC, Normandie Univ\\
    \textsuperscript{\rm 3}GESIS Leibniz Institute for the Social Sciences\\
    firstAuthor@usc.es
}

\begin{document}

\maketitle

\begin{abstract}
The spread of fake news, polarizing, politically biased, and harmful content on online platforms has been a serious concern. With large language models becoming a promising approach, however, no study has properly benchmarked their performance across different models, usage methods, and languages. This study presents a comprehensive overview of different Large Language Models adaptation paradigms for the detection of hyperpartisan and fake news, harmful tweets, and political bias. Our experiments spanned 10 datasets and 5 different languages (English, Spanish, Portuguese, Arabic and Bulgarian), covering both binary and multiclass classification scenarios. We tested different strategies ranging from parameter efficient Fine-Tuning of language models to a variety of different In-Context Learning strategies and prompts. These included zero-shot prompts, codebooks, few-shot (with both randomly-selected and diversely-selected examples using Determinantal Point Processes), and Chain-of-Thought. We discovered that In-Context Learning often underperforms when compared to Fine-Tuning a model. This main finding highlights the importance of Fine-Tuning even smaller models on task-specific settings even when compared to the largest models evaluated in an In-Context Learning setup - in our case LlaMA3.1-8b-Instruct, Mistral-Nemo-Instruct-2407 and Qwen2.5-7B-Instruct. 
\end{abstract}

\begin{links}
    \link{Code and Dataset}{https://github.com/HikariLight/hyperpartisanship_classification/tree/main}
\end{links}

\section{Introduction}

Politically biased (PB), hyperpartisan (HP) and fake news (FN) as well as harmful (HF) social media content when covering divisive topics (e.g. politics, COVID-19) present a significant challenge to public discourse and democratic integrity, and most of those phenomena of our interest can fall under the misinformation category \cite{Wardle}. FN refers to fabricated stories that mimic legitimate news formats \cite{Lazer2018}. HP, on the other hand, involves misleading coverage of real events presented with a strong partisan bias \cite{potthast_stylometric_2018, systematichp}. Both often contain politically charged messages that distort facts and polarize audiences. PB reporting further complicates the media landscape by subtly influencing and polarizing public opinion and eroding the impartiality expected of journalism \cite{zhou}. While the concept of harmful content (HF) is broad and its content can vary based on political priorities \cite{eyuboglu2023fight}, its diffusion on social media can rely on the spread of hate speech \cite{yang-etal-2023-hare} or misinformation \cite{eyuboglu2023fight}. Indeed, harmful forms like polarizing content are frequently fueled by false or misleading narratives to incite frustration \cite{cinelli2021dynamics, OSMUNDSEN_BOR_VAHLSTRUP_BECHMANN_PETERSEN_2021}. Consequently, the detection of such content necessitates robust fact-checking and verification \cite{CheckThat2022}. This conceptual overlap is reflected in recent research challenges. For example, \citet{azizovOverviewCLEF2023CheckThat} considered harmful tweet detection a subtask of identifying relevant claims in tweets containing COVID-19 information. By defining a tweet as harmful when it potentially contained misinformation on COVID-19, they effectively demonstrated how, in a specific and critical context, the tasks of detecting harmful content and misinformation are intrinsically linked. Accurate detection of these diverse forms of problematic content is crucial. Large Language Models (LLMs) are valuable tools for this task. The dominant approaches have been fine-tuning (FT) both encoder-only models \cite{howard2018universallanguagemodelfinetuning} and decoder-only LLMs \cite{aman2024large}. However, a systematic comparison of these models' performance on the specific tasks of fake and hyperpartisan news, politically biased news, and harmful content detection, especially across multiple languages, is still missing from the literature. Our work fills this gap with a comprehensive comparison of encoder-only and decoder-only LLMs using both fine-tuning (FT) and various in-context learning (ICL) settings. Our ICL methods include: zero-shot prompts with different degrees of task specifications and with rule-based approaches (e.g., codebooks); Few-shot (FS) using both randomly selected examples, and diversity-optimized examples selected through Determinantal Point Process (DPP); Chain-of-Thought (CoT). We conducted experiments on 10 datasets in five different languages, moving beyond the common limitation of using only English or U.S.-centric data \cite{systematichp}. This study addresses three key research questions:
\begin{itemize}
    \item \textbf{RQ1}: How do different LLM adaptation paradigms (FT vs. ICL) compare for these tasks, considering model architecture, size, and pre-training data?
    \item \textbf{RQ2}: What is the impact of various ICL strategies—including few-shot examples and rule-based methods—on performance and stability?
    \item \textbf{RQ3}: How do performance and optimal strategies for these tasks vary across different languages, especially for mid- and low-resource languages?
\end{itemize}

Our extensive experiments reveal that FT remains a highly effective technique, often outperforming ICL strategies. Specifically, fine-tuned decoders performed better for PB and FN detection, while encoders were more effective for HP and HF tweets. Within ICL, the codebook approach was generally the effective, outperforming CoT for classification. The use of DPP for few-shot example selection sometimes reduced classification variance, though it did not consistently boost performance.

\section{Related Work}\label{RelatedWork}

\paragraph{Fine-tuning and Political Text Classification}\label{TextClassification}
Text classification is an NLP task that assigns a label to a given text. FT adapts a pre‑trained model to a specific task by further training the model together with a newly added classification head using task‑specific labeled data. \cite{howard2018universallanguagemodelfinetuning}. While effective, this process can be sensitive to the classifier head's initialization \cite{yang-etal-2022-parameter}. In political text classification, FT has been successfully applied to various tasks. For instance, \citet{politics} fine-tuned RoBERTa to create POLITICS, achieving state-of-the-art performance on SemEval 2019 for hyperpartisan news detection. Other works have explored stylistic features for hyperpartisan content discrimination \cite{potthast_stylometric_2018}, created new datasets for multiclass hyperpartisan detection using fine-tuned BERT models \cite{lyu_computational_2023}, or combined BERT with ELMo to enhance FT \cite{naredla_detection_2022}. More recently, LLMs like Llama 2 have been fine-tuned for tasks such as FN detection, leveraging their understanding and analytical capabilities \cite{aman2024large, pavlyshenko2023analysis}.

\paragraph{In-Context Learning}\label{ICL}
With the advent of recent decoder-only LLMs, ICL has emerged as a valuable technique in NLP. Users interact with models directly through prompts, specific textual templates containing instructions and optionally examples \cite{Brown2020LanguageMA}. This approach allows models to perform tasks without prior task-specific fine-tuning \cite{Efrat2020TheTT}, leveraging a single pre-trained model for various downstream tasks and enabling desired behavior specification via natural language. ICL has shown remarkable performance on challenging reasoning tasks \cite{Brown2020LanguageMA, Wei2022ChainOT}. However, ICL is highly sensitive to input format and order \cite{lu-etal-2022-fantastically, min-etal-2022-rethinking}, and can lead to irreproducible outcomes as slight prompt changes significantly impact performance \cite{lu-etal-2024-prompts, sun2023evaluating}. To overcome these limitations, our work comprehensively tested various prompt strategies and demonstration selection methods, including DPP for more stable performance, and introduced codebook prompting. Research on prompt design aims to elicit better performance and reasoning. Notable approaches include CoT prompting \cite{Kojima2022LargeLM}, which encourages step-by-step reasoning, and its zero-shot variant \cite{Wei2022ChainOT}. \citet{lu-etal-2022-fantastically} also highlighted the importance of careful prompt format and example selection in few-shot learning. Regarding the comparison of ICL and fine-tuning, \citet{labrak-etal-2024-zero-shot} and \citet{edwards-camacho-collados-2024-language-models} demonstrated that fine-tuning smaller models often outperforms ICL in larger language models across various NLP and text classification tasks. Aligned with these findings, our study expands this comparison by evaluating a broader range of prompt strategies, models (including ModernBERT \cite{warner2024smarterbetterfasterlonger} beyond Llama models), and by focusing on specialized domains to rigorously compare the efficacy of these tools.

\paragraph{Codebook}\label{Codebook}
A codebook provides definitions of categories, including examples and classifying instructions. For instance, \citet{vincent} developed a codebook to classify hyperpartisan news on a 5-point scale, and \citet{ijerph18147556} crafted one for content analysis of COVID-19 articles, classifying tropes and rhetorical strategies to detect misinformation. Codebooks offer a method to explicitly prompt LLMs with structured contexts, eliciting their rule-based reasoning capabilities, which goes beyond standard ICL that often relies primarily on examples. \citet{hu-etal-2024-leveraging} explored codebook application in zero-shot settings with closed models for political phenomena classification, using the codebook as a structured framework for interpretation. Similarly, \citet{halterman2025codebookllmsevaluatingllms} utilized codebooks to evaluate open models, demonstrating how these guidelines facilitate assessing an LLM's adherence to predefined classification logic. While these specific codebooks were tailored to different political phenomena, NLP tasks, or modeled tasks differently from our dataset intentions, thus not directly applicable to our study, they still served as valuable inspiration for our experimental setup and underscore the potential of integrating structured rule sets into LLM prompts. This enhances adherence to specific classification schemes, aligning with broader prompt engineering efforts to elicit precise and controlled LLM outputs, especially in domains requiring nuanced categorization criteria.

\paragraph{Misinformation and Bias Detection}
LLMs demonstrate reasoning capabilities across various applications, including misinformation detection \cite{li-etal-2023-chatgpt, leite2025weakly}. However, LLMs pose a dual challenge: they can be misused to spread misinformation, and their detection capabilities may diminish with implicit or newly crafted content \cite{chen2024combating}. Consequently, the efficacy of detection methods relative to the rapid updating rate of misinformation is a significant concern \cite{jiang2024disinformation}. In misinformation studies, several works have explored LLM capabilities. \citet{jose2024large} compared proprietary models (GPT, Claude) for zero-shot propaganda detection, finding their performance inferior to RoBERTa-CRF. For hyperpartisan detection, \citet{magginiLeveragingAdvancedPrompting} showed that increased prompt complexity and external knowledge usually improved Llama-3.1-8b-Instruct's performance. Conversely, \citet{omidi-shayegan-etal-2024-evaluation} found encoder models like RoBERTa generally outperformed generative LLMs (GPT-3.5) for Persian hyperpartisan content. In Fake News Detection, \citet{anirudh2023multilingual} observed gpt-3.5-turbo's superiority over a bi-directional transformer for Tamil classification. Notably, while these works explore various aspects, none of them have focused on a comprehensive benchmarking across different ICL strategies, models, and multilingual contexts, which is a key contribution of our study.

\section{Experimental Setting}\label{ExperimentalSetting}

\subsection{Task Formulation}
The core objective of this study is to evaluate the effectiveness of various NLP models in HP, FN, PB and HF detection, by comparing two widely used approaches: FT and ICL, where models are tested as off-the-shelf tools without additional training. Specifically, we focus on tasks such as identifying hyperpartisan and fake news, harmful tweets and a news' political leaning, recognizing the distinct linguistic and contextual challenges each task presents. Our approach encompasses both binary and multi-class classification scenarios, leveraging a variety of datasets and multilingual contexts. While the ultimate goal is not to develop production-ready models, we prioritize thorough experimentation with various transformer-based architectures, prompt strategies, and learning techniques. This exploration serves to highlight the strengths and limitations of the tested architectures, contributing to the broader effort of refining misinformation detection methodologies within the NLP community. We acknowledge the fact that hyperpartisan news shows peculiar stylistic traits, rather than fake news \cite{potthast_stylometric_2018}.

\subsection{Datasets}\label{Dataset}
For our experiments, we selected datasets for both binary and multiclass classification tasks. The 10 datasets focus on articles, headlines, tweets on COVID-19 and political news and different topics including TV, politics, sports, and health. They cover two types of domains: news and Twitter, and include four specific-oriented classification tasks: hyperpartisan and fake news, harmful tweet and political bias detection. For hyperpartisan detection, we selected the \textbf{SemEval-2019 Task 4 by-article} dataset \cite{Kiesel2019SemEval2019T4}, which contains articles from hyperpartisan and mainstream websites annotated by three annotators. They mostly cover the first term of Trump, Gun Control and other U.S.-centric related topics. The dataset's strength lies in its article-level annotations that allow for analysis of extended argumentative structures and narrative techniques typical of hyperpartisan content, rather than just isolated claims. The \textbf{VIStA-H} dataset \cite{lyu_computational_2023} includes hyperpartisan and neutral news headlines from right, center and left U.S. newspapers. Its focus on headlines rather than full articles complements the SemEval dataset. The dataset's temporal coverage, from 2014 to 2023, is particularly valuable for tracking how hyperpartisan news evolved through multiple election cycles.
The \textbf{Fake News Net} dataset \cite{Shu_FNNet} contains news articles shared on Twitter, allowing for analysis of how hyperpartisan content circulates in social media environments. 
The \textbf{Spanish Fake News Corpus} \cite{gomez2021overview} gathers news from Spanish newspapers and media company websites, along with fact-checking websites. This corpus was selected to broaden the linguistic and cultural scope of the research beyond English-language and U.S.-centric media. The incorporation of fact-checking websites provides an additional layer of verification that strengthens the dataset's reliability.
The \textbf{Fake.br Corpus} \cite{fakebr} focuses on Brazilian Portuguese manually collected and checked news. The inclusion of this Brazilian Portuguese corpus further expands the cross-cultural and multilingual dimensions of the research. Brazil's distinct political landscape and media environment provide an important comparative case for testing the generalizability of the detection approaches. The manual verification process strengthens the dataset's reliability as a benchmark for testing detection methods in a language where NLP resources might be less abundant than for English or Spanish. \textbf{CLEF 2022 CheckThat! Lab Subtask 1C} \cite{CheckThat2022} was selected for its focus on COVID-19 misinformation tweets across multiple languages (Arabic, Bulgarian, and English), providing an opportunity to study fake news content around a global crisis that transcended national boundaries. Regarding political bias detection, we considered the \textbf{Qbias} dataset \cite{Qbias}, which contains articles from AllSides, a news aggregator with an established methodology for evaluating political leaning. This provides a more nuanced and professionally curated ground truth for political bias than many other available resources. This nuanced approach helps move beyond binary classifications of political content and supports more sophisticated analysis of bias indicators. Lastly, \textbf{CLEF 2023 CheckThat! Lab Task 3A} dataset \cite{azizovOverviewCLEF2023CheckThat} provides contemporary examples that reflect the current state of political communication and media bias. This diverse collection of datasets provides a comprehensive foundation for developing and evaluating models across multiple languages, cultural contexts, media formats and misinformation tasks. The URLs to retrieve the datasets can be found in the Appendix. To produce input for the classifier in the Spanish Fake News Corpus and Qbias datasets, we concatenated the headline and the body of the article. The datasets are summarized in Table \ref{tab:datasets}.

\begin{table*}[ht!]
\centering
\resizebox{\textwidth}{!}{%
\begin{tabular}{p{6cm}|l|l|l|r|r|r|r|l|l|l|p{5.8cm}}
\textbf{Dataset} & \textbf{Abbr.} & \textbf{Lang.} & \textbf{Timeframe} & \textbf{Train Size} & \textbf{Test Size} & \textbf{Avg. Tkn Train Len.} & \textbf{Avg. Tkn Test Len.} & \textbf{Domain} & \textbf{Type} & \textbf{Task} & \textbf{Label Ratios (Train / Test)} \\
VIStA-H \cite{lyu_computational_2023} & HV & en & 2014–2023 & 1999 & 201 & 13 & 13 & News & S & HP & HP: 0.39/0.63; N: 0.50/0.50 \\
SemEval-2019 by-article \cite{Kiesel2019SemEval2019T4} &SH & en & 2007–N/A & 645 & 628 & 735 & 757 & News & D & HP & HP: 0.50/0.50; N: 0.50/0.50 \\
Spanish Fake News Corpus \cite{gomez2021overview}&SFN & es & 2020–2021 & 676 & 572 & 607 & 843 & News & D & FN & T: 0.50/0.50; F: 0.50/0.50 \\
Fake News Net \cite{Shu_FNNet}& FNN & en & N/A & 18556 & 4640 & 17 & 16 & News & D & FN & T: 0.74/0.74; F: 0.26/0.26 \\
Fake.br Corpus \cite{fakebr}& FBC & pt & 2016–2018 & 5760 & 1440 & 688 & 698 & News & D & FN & T: 0.50/0.50; F: 0.50/0.50 \\ \hline
CLEF 2022 1C \cite{CheckThat2022} &C1A & ar & N/A & 3624 & 1201 & 73 & 68 & Twitter & S & HT & NH: 0.81/0.84; H: 0.19/0.16 \\
& C1B& bu & N/A & 708 & 325 & 62 & 68 & Twitter & S & HT & NH: 0.87/0.97; H: 0.13/0.03 \\
& C1E& en & N/A & 3323 & 251 & 60 & 51 & Twitter & S & HT & NH: 0.91/0.84; H: 0.09/0.16 \\ \hline
CLEF 2023 3A \cite{azizovOverviewCLEF2023CheckThat} & C3A & en & N/A & 45066 & 5198 & 90 & 110 & News & D & PB & R: 0.39/0.13; C: 0.34/0.38; L: 0.27/0.50 \\
Qbias \cite{Qbias} & QB & en & 2012–2022 & 17403 & 4351 & 97 & 96 & News & D & PB & R: 0.39/0.13; C: 0.34/0.38; L: 0.27/0.50 \\
\hline
\end{tabular}
}
\caption{Description of datasets used in our experiments. Average token length (Train/Test) is computed with the Llama3.1-8b tokenizer. Dataset types: S = Sentence, D = Document. Tasks: HP = Hyperpartisan News Detection; FN = Fake News Detection; HT = Harmful Tweet Detection; PB = Political Bias Detection. Labels: HP = Hyperpartisan, N = Neutral; T/F = True/Fake; NH/H = Non-harmful/Harmful; R/C/L = Right/Center/Left. Abbr. contains the abbreviation we will use in this paper to refer to the datasets.}
\label{tab:datasets}
\end{table*}

\subsection{Models}
For our experiment, we compared two types of model architectures: encoder-only and decoder-only models.  

\textbf{Encoder-Only Masked Language Models: } Following \citet{edwards-camacho-collados-2024-language-models}, we selected BERT-derived models such as RoBERTa-base (125 million parameters) and RoBERTa-large (354 million parameters) \cite{liu2019roberta}, XLM-RoBERTa \cite{conneau2019unsupervised}, POLITICS \cite{politics}, which has been adapted for the political domain using continuous pre-training, and mDeBERTaV3 \cite{he2021debertav3}. RoBERTa is pre-trained on English data, while XLM-RoBERTa is trained on 100 different languages, making it suitable for evaluating the impact of multilingual training on performance. 

\textbf{Decoder-Only Large Language Models:} 
We experiment using different small-size open-weight LLMs: LlaMA3.1-8B and LlaMA3.1-8B-Instruct, Mistral-Nemo-Instruct-2407 \cite{mistralnemo}, Qwen2.5-7B-Instruct \cite{qwen2025qwen25technicalreport}. These models are decoder-only and testing them allows us for generalizable effects across model families and tasks. The temperature was set to 0 for all the experiments. Generally, for non-English datasets, we evaluated models exclusively trained on multilingual datasets to ensure appropriate language coverage and performance. 
Further details are given in the Appendix \ref{sec:CIH}.

\subsection{Prompt design}\label{sec:prompt}
Earlier studies like \cite{Wei2022ChainOT}, \cite{jung-etal-2022-maieutic} and \cite{mishra-etal-2022-cross} have demonstrated the effectiveness of using task-specific prompts. Therefore, following \cite{edwards-camacho-collados-2024-language-models} and \cite{labrak-etal-2024-zero-shot}, we constructed the prompts concatenating the following elements: 1) an instruction detailing the task, domain, and describing the meaning of the label; 2) the input argument, supplying essential information for the task; 3) the constraints on the output space, guiding the model during output generation. To improve the coherence, the specificity of the prompts, the instructions to follow in the codebook, and the fine-grained reasoning in CoT for the political domain, we collaborated with an expert in Political Science. In particular, to structure and develop our codebooks, we were inspired by \citet{vincent} and \cite{halterman2025codebookllmsevaluatingllms}, which introduced clear task definitions, explicit and exhaustive rules to determine the label of a data point, as well as provided examples covering both correct and borderline cases. We tested different prompting and ICL strategies such as zero-shot, Few-Shot, codebook and a variant of guided CoT, intending the reasoning as a multi-task evaluation \cite{lee-etal-2024-small-language, duan-etal-2024-reta}, to provide explainable results like in \cite{yang-etal-2023-hare}. We compare the random selection of Few-Shot exemplars with a more diversified selection using DPP, for which we provide a brief introduction in the following section. We also compare the results given by prompting the models with instructions containing different levels of complexity: general instructions, specific definitions of political phenomena or specialized instructions with more context provided. During the prompt optimization phase, we placed particular emphasis on ensuring that the model adhered to a consistent label format. This was crucial to ensure the outputs were reliably parseable. For instance, we discovered that the models when asked to provide string labels generated few unparseable outputs, considered wrong at the time of inference. This behavior happened across all the models and configurations. Moreover, at the begininnig we tested Mistral-7B and most of the time it did not follow the instructions regarding the template. This is why we did not introduce it in the experimental setting. Lastly, in CoT, to ensure its right functioning, we made sure the models generated the thoughts before the final output. Please, see the Appendix for further details. 
%The pipeline of the experiment is shown in Figure~\ref{fig:pipeline}. 

\subsection{Determinantal Point Process} \label{subsec: dpp}
Determinantal Point Process is a probability distribution over cloud of points that are used as computational tools across the fields of physics, statistics and machine learning \cite{gautieretal2019}. DPP has been used to select diverse and representative set of datapoints for in-context learning \cite{yang-etal-2023-representative}, data annotation \cite{wang2024effectivedemonstrationannotationincontext}, instruction tuning \cite{wang2024diversitymeasurementsubsetselection} and pre-training \cite{yang2024p3policydrivenpaceadaptivediversitypromoted}. DPP has been preferred for these tasks because it helps in promoting efficiency while maintaining a diversity of the selected subset from a large set. For this let us take 2 sets called index set $A =\{1, 2,\dots N\}$ and its corresponding item set $I_{A} = \{x_{1}, x_{2}, \dots, x_{N}\}$. Then the problem of subset selection becomes evaluating $2^{M}$ subsets, which is computationally intractable and combinatorially explosive as the size of the super set grows. In order to approximately solve this problem DPP first uses the representation of the data $\mathbf{x_{i}}$. We use Sentence-BERT \cite{reimers-gurevych-2019-sentence} to calculate the representations or embeddings. After the representations are computed, DPP algorithm works by calculating a Kernel $\mathbf{K}_{ij} = k(\mathbf{x_{i}}, \mathbf{x_{j}})$ where kernel function k can be any similarity or distance metric between 2 points. Based on this we select a subset $Y \in A$, the probability selection of Y is given by.
\begin{equation}
 P(Y) = \frac{det(\mathbf{K}_{Y})}{det(\mathbf{K} + \mathbf{I})}
\end{equation}
Here $\mathbf{K}_{Y}$ is the subset of the matrix $\mathbf{K}$ and consists of $\mathbf{K}_{ij}$ for $i,j \in Y$. $\mathbf{I}$ is the identity matrix and $det(\cdot)$ represents the determinant of a matrix. Under these conditions the selection of the best subset can be formulated as the optimization problem as follows:

\begin{equation}
 Y_{best} = argmax_{Y \subset A, |Y| = k} det(L_{Y})
\end{equation}
Algorithms exist to select such subsets of size k from the superset by sampling from the posterior distribution. For the task of selection we use the exact sampler \cite{gautieretal2019, mazoyer2020projectionsdeterminantalpointprocesses} which is the part of DPPy package by \cite{gautieretal2019} and is faster than Monte-Carlo sampler \cite{bardenet2019montecarlodeterminantalpoint}. Since the algorithm works on sampling at the kernel level it helps in selecting datapoints which are diverse in the representation space. %For more theoretical guarantees readers can refer to \cite{pmlr-v80-celis18a}.

\subsection{ICL Setting}
Our aim is to compare the capabilities of different learning techniques, namely FT and ICL, and model architectures for hyperpartisan, fake news, harmful tweet and political bias classification. 
To investigate the ability of LLMs on those tasks in ICL, we used LlaMA3.1-8b-Instruct, Mistral-Nemo-Instruct-2407 and Qwen2.5-7B-Instruct by prompting them with different setups: 0-shot with General Prompt, 0-shot with Specific Prompt, CoT and Few-Shot with \textit{k}-shot where \textit{k}>0. In the \textit{k}-shot configuration, we adopted the General Prompt along with random examples and the respective labels of the dataset. To further test the stability of the prompting, we used Determinantal Point Process \cite{gautieretal2019} to select a diverse set of datapoints for each of the k-shot settings. The general prompt template is: $<$Role$>$$<$Task description$>$$<$Definition or Instructions$>$$<$Text to classify \textit{or} examples followed by text to classify$>$$<$Response format$>$. We provide examples of different our prompts in the Appendix (see Tables 4, 5, 6 and 7). In order to maintain a balanced pool of examples, for multi-class datasets, we sampled from 1 to 3 examples per label; otherwise, we extracted the same \textit{k}-shot per class. Finally, for non-English data, the corresponding roles and instructions were provided in the respective language to ensure accuracy and contextual relevance.
The type of prompts we used are the following:
\textbf{General Prompt} By providing the model with task-specific context (e.g., a headline, article, or tweet), we prompted it to classify the input text with the appropriate task label. With this configuration, we leverage the internal knowledge of the model to predict the answer, while being aware that it can suffer from political bias \cite{Bang2024MeasuringPB}. We used it in 0-shot and Few-Shot.

\textbf{Specific Prompt} We slightly changed the previous template, introducing in the instruction the political definition of the phenomenon analyzed and some knowledge regarding the biases in partisan texts and asked the model to classify the text with the correct label. These political definitions were provided by a domain expert. Thus, we insert external knowledge and introduce a political definition to maximize model's understanding capability and improve its outputs' quality. We tested its efficacy only in zero-shot.

\textbf{Codebook} Based on previous works, with the help of a Political Scientist we crafted a specific codebook for each task. These codebooks contain a definition of the phenomenon, a description of the task's characteristics considering several aspects (e.g., style, narrative) and particular linguistic features (e.g., use of hashtags, tone, source credibility). Furthermore, the detection criteria contain examples to help the LLM in understanding the specific rules for each task. Crucially, this detailed codebook information was directly embedded within the <Definition or Instructions> component of our prompt template, allowing the LLM to perform rule-based reasoning for classification. This approach enables the models to leverage explicit, structured knowledge during inference, directly addressing the complexities of the tasks.

\textbf{Guided CoT Prompt} We guided the model to break down its reasoning step by step before making a final classification on the specified context, ensuring it produced explanations for all the steps before the final prediction. Specifically, we divided the hyperpartisan classification task into different sub-tasks: sentiment analysis, rhetorical bias, framing bias, ideology detection \cite{magginiLeveragingAdvancedPrompting}. Moreover, by asking the model to identify itself with a specific political leaning, we are introducing recursive thinking \cite{duan-etal-2024-reta}. This method encourages the model to consider multiple factors and clearly articulate its reasoning, potentially leading to more robust and explainable classifications. Lastly, our approach allows us to consider the multidimensionality of each misinformation phenomenon analyzed. 

By guiding the model through this structured reasoning process, we aimed to reduce misclassification and promote a more nuanced analysis. This approach also enabled us to observe how the model weighs different textual elements in its decision-making process, which helps in identifying any inherent biases or limitations within the model's reasoning. We conducted preliminary tests with various prompts and configurations to refine the ones used in this experiment, ultimately selecting the configurations that yielded the best results on the training set. The optimization of these prompts was done manually rather than through automated methods. As a result, our prompts vary in terms of length, complexity, task specificity, and domain relevance, providing a comprehensive range of settings for evaluation. This structured and manually-optimized approach not only enhances the model's classification performance but also provides deeper insights into the model's interpretability and decision-making process across different political contexts.

\begin{table*}[h!]
\centering
\resizebox{\linewidth}{!}{%
\begin{tabular}{|l|c|c|c|c||c|c|c|c||c|c|c|c||c|c|c||}
\hline
\textbf{Model} &  & \textbf{HV} & \textbf{SH} & \textbf{Macro Avg.\ HP}
                & \textbf{FNN} & \textbf{SFN} & \textbf{FB} & \textbf{Macro Avg.\ FN}
                & \textbf{C1A} & \textbf{C1B} & \textbf{C1E} & \textbf{Macro Avg.\ HF}
                & \textbf{QB} & \textbf{C3A} & \textbf{Macro Avg.\ PL} \\ \hline

\multirow{2}{*}{RoBERTa-base}
  & Acc & .822 & \textbf{.865} & .843 & .879 & –    & –    & –    & –    & –    & .919 & –    & .622 & .659 & .640 \\
  & F1  & .818 & \textbf{.865} & .841 & .880 & –    & –    & –    & –    & –    & .915 & –    & .604 & .660 & .632 \\ \hline

\multirow{2}{*}{RoBERTa-large}
  & Acc & \textbf{.852} & \textbf{.865} & \textbf{.858} & .893 & –    & –    & –    & –    & –    & \textbf{.925} & –    & .683 & .663 & .673 \\
  & F1  & \textbf{.850} & \textbf{.865} & \textbf{.857} & .893 & –    & –    & –    & –    & –    & \textbf{.923} & –    & .674 & .660 & .667 \\ \hline

\multirow{2}{*}{XLM-RoBERTa}
  & Acc & .827 & .801 & .814 & .842 & .623 & .957 & .807 & \textbf{.917} & .917 & .917 & \textbf{.917} & .582 & .632 & .607 \\
  & F1  & .825 & .798 & .811 & .844 & .577 & .957 & .793 & \textbf{.883} & .884 & .885 & \textbf{.884} & .553 & .629 & .591 \\ \hline

\multirow{2}{*}{POLITICS}
  & Acc & .831 & .854 & .842 & .867 & –    & –    & –    & –    & –    & .916 & –    & .682 & .679 & .680 \\
  & F1  & .826 & .854 & .840 & .868 & –    & –    & –    & –    & –    & .911 & –    & .673 & .678 & .675 \\ \hline

\multirow{2}{*}{mDeBERTaV3}
  & Acc & .819 & .776 & .797 & .893 & –    & –    & –    & –    & –    & .915 & –    & .539 & .590 & .564 \\
  & F1  & .816 & .772 & .794 & .841 & –    & –    & –    & –    & –    & .874 & –    & .534 & .570 & .552 \\ \hline

\multirow{2}{*}{ModernBERT-large}
  & Acc & .829 & .854 & .839 & .858 & \textbf{.863} & .941 & .883 & .815 & .965 & .835 & .872 & .658 & .654 & .653 \\
  & F1  & .824 & .853 & .839 & .846 & \textbf{.863} & .941 & .815 & .815 & .949 & .803 & –     & .649 & .657  & .667 \\ \hline

\multirow{2}{*}{ModernBERT-base}
  & Acc & .764 & .780 & .767 & .852 & .782 & .942 & .854 & .830 & \textbf{.966} & .830 & .875 & .584 & .617 & .592 \\
  & F1  & .755 & .779 & .767 & .840 & .781 & .942 & .854 & .792 & \textbf{.966} & .774 & .844 & .571 & .612 & .592 \\ \hline\hline

\multirow{2}{*}{LlaMA3.1-8b}
  & Acc & .830 & .810 & .820 & \textbf{.945} & .812 & .975 & \textbf{.911} & .864 & .869 & .921 & .884 & .788 & .762 & .775 \\
  & F1  & .830 & .801 & .815 & \textbf{.945} & .801 & .975 & \textbf{.907} & .858 & .832 & .920 & .870 & .786 & .763 & .774 \\ \hline

\multirow{2}{*}{LlaMA3.1-8b-Instruct}
  & Acc & .784 & .820 & .801 & .875 & .823 & \textbf{.976} & .890 & .878 & .915 & .862 & .885 & .786 & .796 & .791 \\
  & F1  & .782 & .820 & .801 & .869 & .823 & \textbf{.976} & .889 & .867 & .928 & .829 & .875 & .781 & \textbf{.802} & .792 \\ \hline

\multirow{2}{*}{Mistral-Nemo-Instruct-2407}
  & Acc & .834 & .736 & .783 & .867 & .725 & \textbf{.976} & .851 & .850 & .943 & .838 & .877 & \textbf{.790} & \textbf{.846} & \textbf{.819} \\
  & F1  & .833 & .733 & .783 & .855 & .722 & \textbf{.976} & .851 & .859 & .946 & .825 & .877 & \textbf{.787} & .789  & \textbf{.849} \\ \hline

\multirow{2}{*}{Qwen2.5-7B-Instruct}
  & Acc & .819 & .686 & .745 & .864 & .685 & .974 & .835 & .856 & .913 & .824 & .864 & .735 & .698 & .711 \\
  & F1  & .812 & .677 & .745 & .855 & .675 & .974 & .835 & .863 & .928 & .806 & .866 & .728 & .693 & .711 \\ \hline

\end{tabular}%
}
\caption{Performance of models in the FT setting. The reported weighted Accuracy and weighted F1 scores are the averages obtained by running each model five times on the same dataset, reporting standard deviation.}
\label{tab:FT_results}
\end{table*}

\section{Main Results and Discussion}

\subsection{Fine-Tuning}
Table \ref{tab:FT_results} presents the results for fine-tuning. On average across datasets, decoder-based models tend to outperform encoder-based models on tasks that require factual world knowledge, such as fake news detection and political bias identification. For instance, in fake news detection (Macro Avg. FN F1 score), the LlaMA3.1-8b (decoder) achieves .907, while the best performing encoder with a directly comparable macro average, ModernBERT-base, scores .854. Similarly, for political leaning detection (Macro Avg. PL F1 score), the Mistral-Nemo-Instruct (decoder) reaches .849, significantly surpassing the top encoder, POLITICS, which scores .675. Conversely, encoders achieve better results on linguistically oriented tasks, specifically harmful tweet detection and hyperpartisan language identification. RoBERTa-large (encoder) records an F1 score of .850 on HV, compared to the best decoder, Mistral-Nemo-Instruct, at .833. For SH, RoBERTa-large again leads with an F1 score of .865, while the best performing decoder on this task, LlaMA3.1-8b-Instruct, achieves .820. We hypothesize that this difference arises because the bidirectional attention mechanism of encoders may be better at capturing nuanced linguistic features, whereas decoders might excel at tasks more reliant on content or semantic understanding. Surprisingly, continuous pretraining of RoBERTa-base aimed at adapting it to either the political domain (POLITICS) or multilingual contexts (XLM-RoBERTa) has, in several instances, not led to improved performance over the original RoBERTa-base model and sometimes resulted in a reduction. For example, on the Macro Avg. HP (Hyperpartisan) task, RoBERTa-base achieves an F1 score of .841, whereas POLITICS scores .840 and XLM-RoBERTa scores .811. Regarding decoder models, we observe that a larger parameter count or expanded training corpus does not necessarily equate to superior results. This is demonstrated by LlaMA 3.1-8B outperforming Mistral Nemo-Instruct-2407 in hyperpartisan detection (SH F1 score of .801 for LlaMA 3.1-8b versus .733 for Mistral Nemo). Meanwhile, the decoder models exhibit relatively comparable high performance in harmful text detection (Macro Avg. HF F1 scores): LlaMA3.1-8b (.870), LlaMA3.1-8b-Instruct (.875), and Mistral-Nemo-Instruct-2407 (.877). During the FT experiment, we reached the SOTA in HV, FNN, SFN and FB. We provide the comparison with the previous research in Table 9 in the Appendix.

\subsection{In-Context Learning}
\begin{figure*}[ht]
\centering
\includegraphics[width=\textwidth]{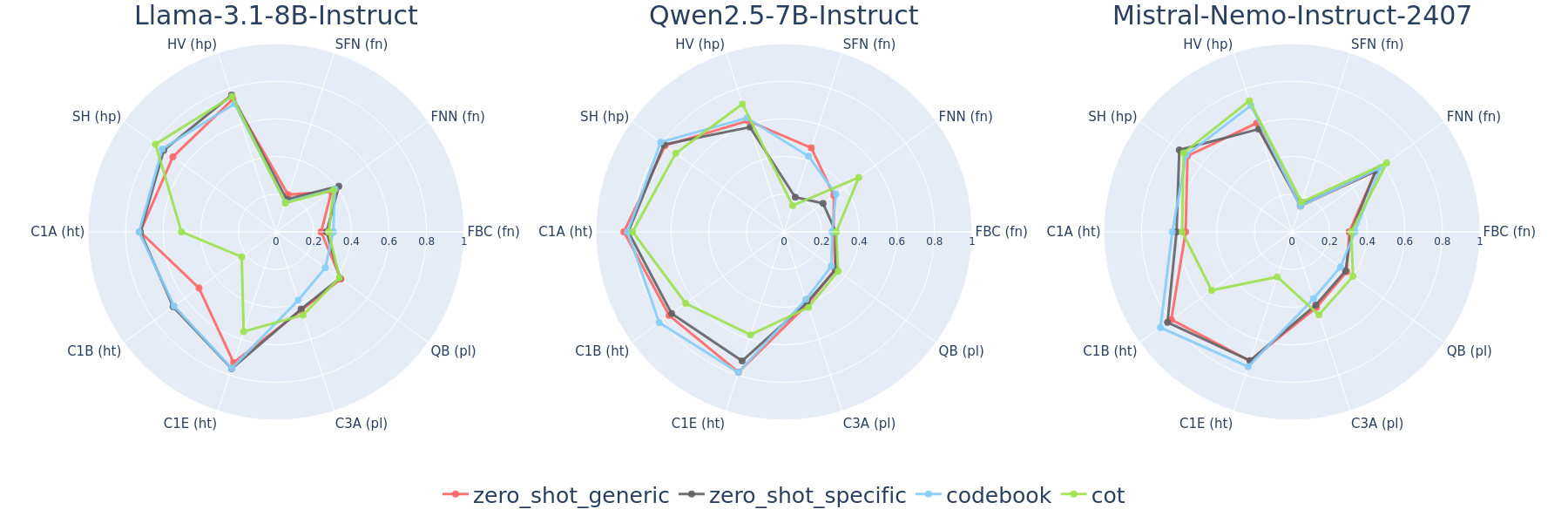} 
\caption{Results for zero-shot and CoT grouped by models.}
\label{fig:Radar1}
\end{figure*}

\begin{figure*}[ht]
\centering
\includegraphics[width=\textwidth]{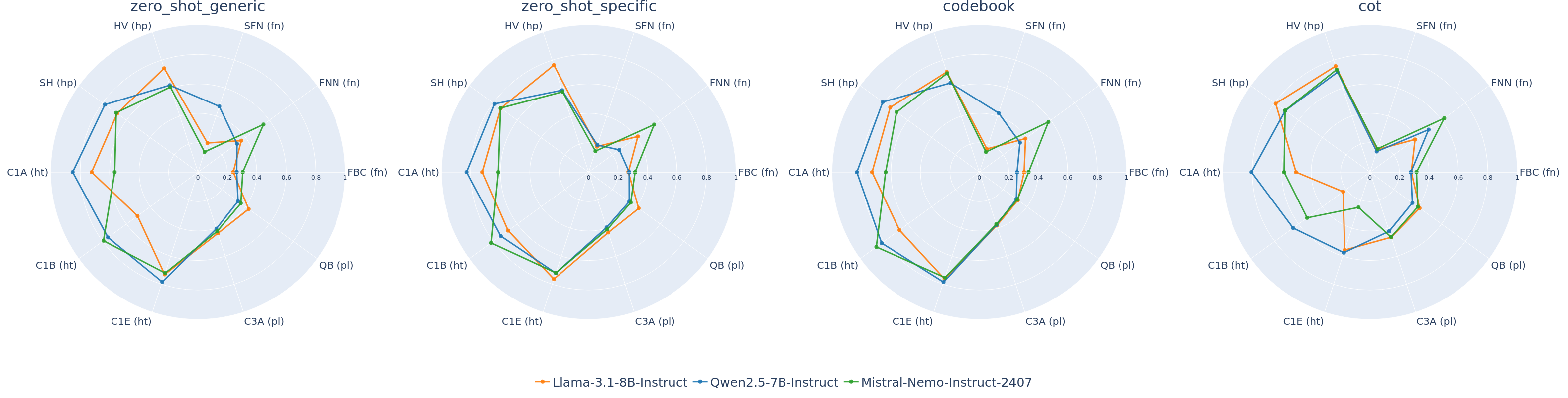} 
\caption{Results zero-shot and CoT grouped by configuration.}
\label{fig:Radar2}
\end{figure*}

For the results discussed in this section, please refer to Figures \ref{fig:Radar1} and \ref{fig:Radar2} for the zero-shot configurations and CoT; and \ref{fig:FS_results} for FS. Detailed results are reported in Table 8 in the Appendix \ref{app:results_SOTA}

\subsubsection{HP}
\par\textbf{zero-shot-general vs zero-shot-specific} In the hyperpartisan (HP) task, moving from generic to specific zero-shot prompting results in moderate improvements across all three models, especially for LLaMA 3.1-8B Instruct and Mistral-Nemo-Instruct, whose F1 scores increase from 0.678 and 0.686 to 0.738 and 0.740, respectively. This performance gap indicates that these models have learned robust representations of hyperpartisan content during pre-training, and that providing more detailed task descriptions helps further refine their predictions. Lastly, SH predictions are generally stronger, largely because this dataset includes full articles rather than just headlines, as in HV. The richer contextual information in SH provides models with more linguistic and semantic cues, enabling a deeper understanding of the content and improving their ability to detect hyperpartisan narratives. In contrast, the limited context in headlines offers fewer signals for accurate classification.
\par\textbf{{Codebook}} The codebook approach yields improvements for SH, where Llama and Qwen demonstrate the most significant gains, particularly on the SH dataset, where their performance approaches F1 .810 and .748 respectively. This suggests that providing explicit criteria for identifying partisan language can help address edge cases where the models depend solely on their internal task representations. The performance gap between models narrows with codebook prompting, indicating that structured guidance can help equalize performance differences stemming from model's architecture, that may rely on different definitions of the phenomenon investigated.

\subsubsection{FN} 
\par\textbf{{zero-shot-general vs zero-shot-specific}} Fake news detection tasks show variable performance under zero-shot conditions. The FNN task proves more challenging are the multilingual datasets: SFN and FBC. The transition from generic to specific prompting yields minimal gains, with Mistral-Nemo-Instruct-2407 maintaining the highest performance on 2 out of 3 fake news (FN) datasets. We observed a 0.275-point drop in F1 score for Qwen on the Spanish dataset, which may be attributed to a misalignment between the fake news definitions—specifically, the model’s internally assumed definition in the zero-shot generic prompt versus the expert-crafted definition used in the zero-shot specific prompt.  This limited improvement suggests that, beyond clear definitions, models need additional contextual or world knowledge to effectively differentiate factual from fabricated content.
\par\textbf{{Codebook}} The codebook approach yields minimal improvements for FNN and SFN. indicates that providing explicit criteria for evaluating factual claims, source credibility markers, and stylistic indicators of fabricated content may help models overcome the inherent complexity of fact verification. The codebook's little effectiveness in this domain suggests that FN does not completely benefit from structured evaluation frameworks. 
Regarding FN, with the different prompt strategies tested in ICL, we found out that the small LLMs are not effectively capable of detecting this kind of disinformation because they can not rely properly on the ontological structures encoded in their world-knowledge. Thus, employing these models as out-of-the-box tools with an ICL setup proves to be inefficient for this task.

\subsubsection{PL}
\par\textbf{{zero-shot-general vs zero-shot-specific}}
Political leaning (PL) classification tasks exhibit low performance across all models under zero-shot conditions. It is important to note that this is a multiclass classification task, which adds complexity. Generally, specific prompting produces a slight decrease for this task. Llama maintains consistently higher performance (on average F1 .416) than the other models. The limited effectiveness with a specific definition of the political wings suggests that PL requires more than definitional refinement to overcome the inherent subjectivity involved. 
\par\textbf{Codebook} Providing more detailed and specific knowledge through the codebook generally resulted in decreased performance across all models. This highlights the complexity of the task and suggests that, despite offering explicit rules to interpret the U.S. political context, agendas, the cultural nuances and linguistic factors involved are insufficient for effectively addressing the task. This result reveals the complexity underlying political bias detection.

\subsubsection{HF}
\par\textbf{{zero-shot-general vs zero-shot-specific}} HF covered three languages and Qwen reached the best results in C1A with zero-shot-generic prompts (F1 .851). However, when prompted with zero-shot-specific prompts, its performance decreased. On the other hand, Llama particularly benefitted from the introduction of the specific knowledge for C1B (from F1 .507 to .676) and C1E (from F1 .730 to .764). 
\par\textbf{{Codebook}} With the introduction of specific dimensions to frame the task, all the models across the HF datasets (except for Qwen in C1A) improved their performances. Specifically, Mistral reached F1 .864 in C1B. This marked improvement highlights the value of explicit harm taxonomies and classification criteria for this sensitive domain. The codebook's effectiveness for harmful content detection suggests that these tasks require clearly articulated boundaries and examples to overcome potential ambiguities in what constitutes harmful material.

\subsection{Summary for Zero-shot configurations}
Across all classification domains, the transition from generic to specific zero-shot prompting lead only to marginal improvements. This pattern suggests that merely elaborating on task definitions provides insufficient guidance for these models to significantly alter their classification behavior. The codebook approach demonstrates modest effectiveness across all task categories, with particularly improvements for HF and FN classification. This pattern indicates that providing structured classification criteria helps models overcome the inherent complexities of these judgment tasks. The codebook's effectiveness stems from its ability to bridge the gap between abstract classification concepts and concrete textual indicators, providing models with clearer decision boundaries for ambiguous cases. Lastly, we noticed that more subjective and nuanced tasks like HF and PL, that require extensive knowledge of the facts and their truthfulness, show improvement with advanced prompting strategies than more stylistic-based tasks (e.g. HP detection), suggesting that prompt optimization benefits may correlate with task complexity. Indeed, the rule based approach is the best ICL configuration for 3 out of 10 datasets: SH, C1B and FBC.

\subsection{FS: DPP-selected vs Random examples}

\begin{figure}[t]
 \includegraphics[width=\columnwidth]{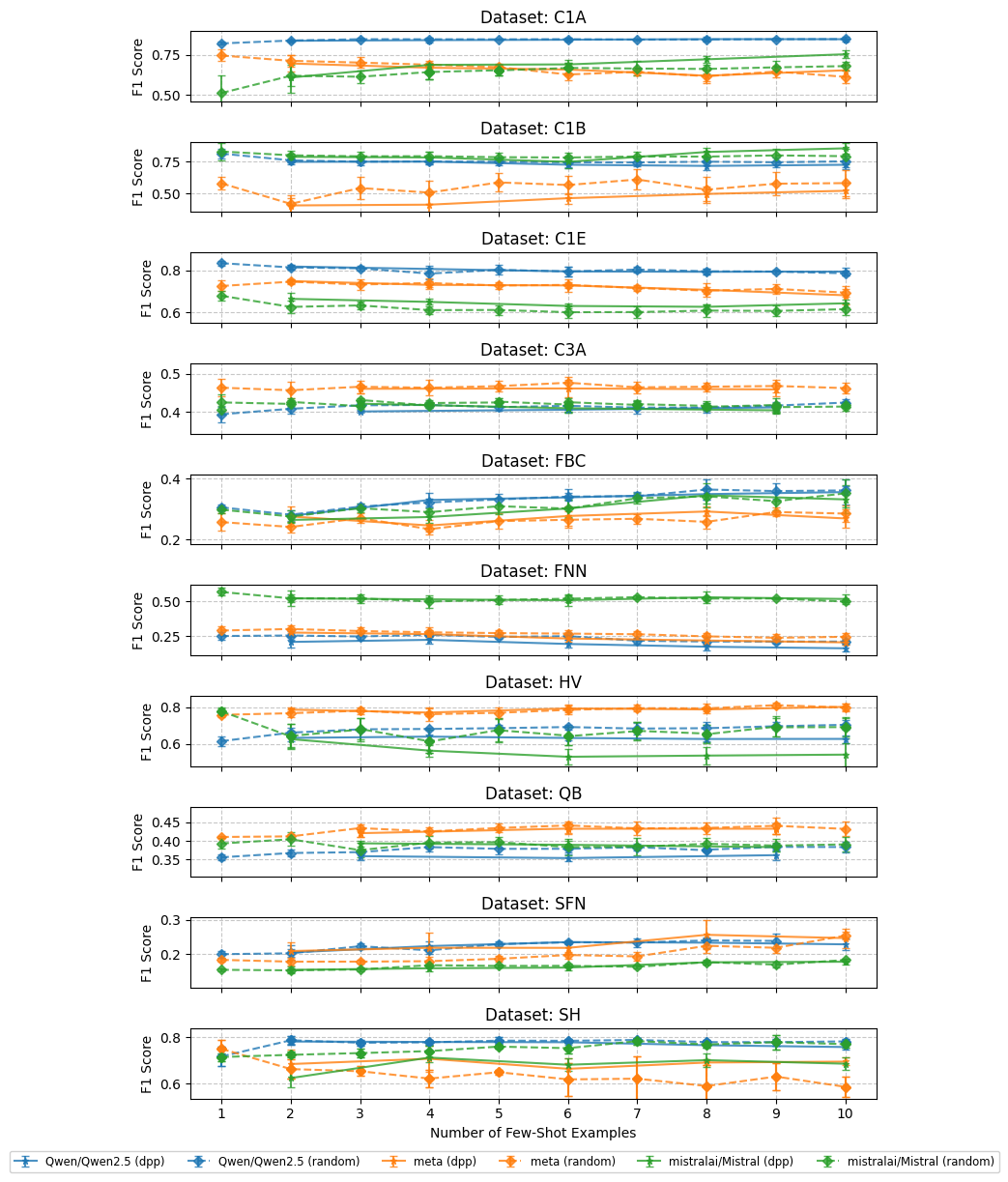}
 \caption{FS DPP vs Random results.}
 \label{fig:FS_results}
\end{figure}

To test the Few-Shot capacity of the model, we decided to compare the performances using random datapoints against a representative set of examples, maintaining the dataset diversity using DPP. Table \ref{fig:FS_results} reports the results for this comparison.
Random selection risks subsets that lack diversity or fail to represent edge cases, while DPP ensures prompt stability by challenging the model with dissimilar patterns. In Few-Shot Learning, where models rely on limited examples to generalize, diverse subsets prevent overfitting to specific features and improve generalization. DPP-selected examples enhance prompt informativeness by showcasing varied cases, enabling the model to better understand nuanced relationships. This diversity reduces classification errors and improves accuracy by covering a broader range of inputs. A key observation across both FS Random and FS DPP is that increasing the number of shots does not consistently or monotonically improve performance for all models and datasets. Generally, performance often peaks at an intermediate number of shots (e.g., 1-shot, 5-shot, 6-shot) and can then plateau, fluctuate, or even decline as more shots are added (e.g., in the range 7-10 shots). This implies that simply providing more examples is not always the best strategy, regardless the datapoints representativeness of the selected examples. There is not any ideal threshold for the n-shots, since the performances vary across models and datasets. 
For instance, while using FS Random examples in Hyperpartisan Detection on the SH dataset, Llama-3.1-8b-Instruct's F1 score varies from .751 (1-shot) down to .623 (4-shot) and then to .587 (10-shot). However, the results do not indicate a clear best method across all scenarios. Peak performance for a given model and dataset can be achieved by either method, often at different n-shot values. For example, with Llama-3.1-8B-Instruct on the HV dataset, Random 9-shot yields an F1 of .811, while DPP 10-shot gives .801. Conversely, Mistral-Nemo-Instruct-2407 on C1B achieves its highest few-shot F1 of .851 with DPP 10-shot. FS Random can achieve high scores, possibly when the random selection happens to include particularly effective examples. However, its performance can be inherently more variable. FS DPP, by design, selects for diversity, which might be expected to lead to more robust or consistent improvements. While it achieves strong results in some cases, it also exhibits fluctuations and doesn't always outperform Random FS.

\subsection{Chain of Thought}
This setting provided reasoning steps with increasing levels of abstraction modeled as successively sub-task steps. Llama-3.1-8B-Instruct shows good performance with CoT on HP tasks, achieving the best F1/Accuracy in its section for both SH (F1 .792, Acc .795) and HV (F1 .757, Acc .764) datasets. It also performs well on PL Detection for C3A (F1 .465, Acc .459) and QB (F1 .416, Acc .405), again leading in its section. Mistral with CoT achieves the overall best scores for FN Detection across all models and configurations on the FNN dataset (F1 .623, Acc .680) and across sections in FBC dataset (F1 .315, Acc .397), and SFN (F1 .167, Acc .168). This suggests CoT might be particularly beneficial for this model on these specific complex tasks. Regarding, Qwen, CoT helps it in achieving strong results on C1A (F1 .804, Acc .783), but only F1 .646 in C1B and F1 .575 in C1E, leading this section for these datasets. Nevertheless, in most of the cases, it revealed to be suboptimal. The CoT prompting proved largely suboptimal across our experiments, showing significant improvement only for the FNN task. Our analysis suggests this underperformance stems primarily from language representation issues in the model's training data. When prompted in underrepresented languages with insufficient training tokens, the models struggled to process the unfamiliar linguistic patterns. Rather than aiding reasoning, these novel tokens appeared to confuse the models, disrupting their inference capabilities. Additionally, we observed that even in zero-shot-specific, codebook, and few-shot configurations, the models sometimes generated explanations unprompted, suggesting they were trained to occasionally provide reasoning alongside their answers. This built-in explanatory behavior likely accounts for why explicit CoT prompting offered minimal additional benefits despite its theoretical advantages and added complexity.

\subsection{Insights from Fine-Tuning vs. In-Context Learning in LLMs}
Across all the configurations tested in our experiments, FT emerged as the most effective method to apply models to the political domain, and, in particular, the disinformation subdomain. Specifically, for LLMs, the FT configuration demonstrated its efficacy in 28 out of 33 cases. Notably, Qwen particularly benefited from ICL achieving its best performance with the following configurations: few-shot random for C1E (F1: 0.833) and SFN (F1: 0.678), and zero-shot codebook for SH (F1: 0.810). These results suggest that updating model parameters through FT is generally the most reliable way to optimize performance for downstream tasks in this domain. However, ICL remains a valid and convenient strategy for probing a model’s task-specific knowledge without parameter updates. Despite our efforts to optimize prompts—by incorporating external domain-specific knowledge, employing rule-based approaches, and eliciting reasoning capabilities—ICL configurations still showed more limited effectiveness compared to FT. Lastly, both model architectures benefited from fine-tuning, with encoder-based models achieving superior performance on 6 out of 10 datasets, and smaller LLMs performing better on the remaining 4—particularly in tasks such as fake news and political leaning detection, which require deeper world knowledge. It is important to note, however, that fine-tuning —especially when applied to LLMs—demands significant computational resources, making it a considerably resource-intensive approach.

\section{Conclusion and Future Work}
This paper provides a comprehensive benchmark for FT and ICL methods across classification tasks in several misinformation domains. Indeed, we largely compared different model architectures, learning techniques and sets of prompts in several classification tasks. We evaluated performance on 10 diverse datasets, spanning binary and multiclass contexts in English, Spanish, Brazilian Portuguese, Arabic, and Bulgarian. Particularly, we compared nine models covering the following transformer family's architecture: (1) encoders: RoBERTa-base, RoBERTa-large, XLM-RoBERTa, POLITICS, Modern-BERT-base and -large, mDeBERTaV3; (2) decoders: LlaMA3.1-8b-Instruct,  Mistral-Nemo-Instruct-2407  and  Qwen2.5-7B-Instruct. ICL consistently proved to be less effective than fine-tuning across most settings. Indeed, results 
showed that in FT, decoders were better at PB and FN detection, whereas encoders were better at HP and HF tweet detection.
Regarding ICL, we applied different levels of prompt optimization, testing all the main ICL techniques. In Few-Shot, we found that DPP sometimes reduces variations in classification rather than randomly sampling the shots, though it does not systematically increase the performance. Furthermore, the adaptation methods in ICL did not behave the same depending on the LLM used. Lastly, except for fake news detection tasks, eliciting the model with a codebook was generally the best approach in ICL, making the CoT unreliable for classification task. 
Future work could investigate the application of a RAG system to incorporate up-to-date information and improve the factual verification of news.

\section{Limitations}\label{Limitations}

\textbf{LLMs}
In CoT, to overcome the different templates generated by the model in the initial phase of our experiment, we crafted a template for the different tasks (see the prompt Tables 4,5,6,7 in the Appendix). Moreover, in few cases the model produced irregular outputs.  When this occurred, we considered those cases to be incorrectly labeled.

\textbf{Practical applications} Our work is a comprehensive overview of architectures and methods. Nevertheless, some datasets do not contain up-to-date information that can be used effectively to tackle fake news propagation, since this kind of misinformation does not rely only on linguistic clues but also on pre-existence knowledge of political facts. Indeed, the temporal limitation can affect how the perception of a fact - a general one - can be perceived and or discussed, because the language is subjected to changes over time. However, those dataset could be used for continual pre-training or as part of a RAG system that also incorporates up-to-date information.

\textbf{Domain} We acknowledge that our experiments focus on a subset of political NLP tasks, specifically misinformation-related tasks across different languages. As such, our findings should not be generalized to the full range of NLP tasks.

\textbf{Open LLMs and size} We limited our model selection to open models, while discarding closed one (e.g. Claude, OpenAI) for the sake of reproducibility and budget limitations. Furthermore, our GPUs could not host larger models. This fact limits our findings.

\section*{Ethics Statement}
One of the primary objectives of this work is to address the challenge of misinformation spreading — a critical issue in today’s society. Tackling misinformation is both beneficial to society and ethically imperative, as it contributes to a more informed and balanced public discourse. However, we acknowledge that our work is not without risks and potential unintended consequences. Our study involves the exploration of various architectures and models to assess their reliability in identifying and countering misinformation. This process inherently carries ethical considerations, particularly related to the datasets we used. The datasets include linguistically hazardous data, such as offensive content in the Harmful Tweet dataset, and highly polarized messages, as seen in the Hyperpartisan News and Political Bias detection datasets. Although the datasets used are crucial for the development of robust and effective models, they also pose risks of misuse. Specifically, such datasets could potentially be exploited to train LLMs capable of generating biased or misleading political content, thereby exacerbating the very problem we aim to mitigate. To address these risks, we took steps to ensure our work adheres to ethical standards. Firsly, regarding data handling, we carefully managed the datasets, using them only for the tests described in our paper and not for other unethical purposes. We emphasize the importance of using the models and methodologies developed in this study exclusively for combating misinformation and promoting ethical information dissemination. Misuse of these tools to create or amplify harmful content is strongly discouraged. Then, by openly releasing our code and datasets, we aim to promote transparency and encourage responsible research practices. We also are going to provide detailed documentation to inform users of the potential risks associated with these datasets. VIStA-H and SemEval-2019, Qbias, and CLEF23 3A contain both headlines and articles with extremely polarized content from both Right and Left wing leanings, slurs and racist sentences. CLEF22 1C ar, bu and en host tweets supporting conspiracy theories related to COVID-19. Spanish Fake News Corpus gather cultural fake news as well as ones against gender equality and the LGBTQIA+ community. Fake News Net and Fake.br-Corpus contain a wide range of fake news on different topics, from political to climate change. Lastly, recognizing the potential for misuse, we encourage future researchers and practitioners to implement safeguards, such as adversarial testing and bias detection frameworks, when deploying these models. We further stress that while our findings demonstrate the superior performance of fine-tuned models over in-context learning strategies, this advantage must be wielded with care. Fine-tuned models can be highly specialized and powerful, making it imperative to ensure they are used responsibly. As researchers, we are committed to fostering a dialogue around the ethical implications of NLP technologies and encouraging their use for the betterment of society.
By highlighting these concerns and promoting transparency, we aim to contribute to a more ethical and responsible approach to NLP research in the context of misinformation detection.

\section*{Acknowledgements}
This project has received funding from the European Union's Horizon Europe research and innovation programme under the Marie Skłodowska-Curie Grant Agreement No. 101073351.

\begingroup
\small
\bibliography{aaai2026}
\endgroup

\section{Ethics Checklist}

\begin{enumerate}

\item For most authors...
\begin{enumerate}
    \item  Would answering this research question advance science without violating social contracts, such as violating privacy norms, perpetuating unfair profiling, exacerbating the socio-economic divide, or implying disrespect to societies or cultures?
    Yes
  \item Do your main claims in the abstract and introduction accurately reflect the paper's contributions and scope?
    Yes
   \item Do you clarify how the proposed methodological approach is appropriate for the claims made? 
    Yes
   \item Do you clarify what are possible artifacts in the data used, given population-specific distributions?
    Not Applicable
  \item Did you describe the limitations of your work?
    Yes
  \item Did you discuss any potential negative societal impacts of your work?
    Yes
      \item Did you discuss any potential misuse of your work?
    Yes
    \item Did you describe steps taken to prevent or mitigate potential negative outcomes of the research, such as data and model documentation, data anonymization, responsible release, access control, and the reproducibility of findings?
    Yes
  \item Have you read the ethics review guidelines and ensured that your paper conforms to them?
    Yes
\end{enumerate}

\item Additionally, if your study involves hypotheses testing...
\begin{enumerate}
  \item Did you clearly state the assumptions underlying all theoretical results?
    Not Applicable
  \item Have you provided justifications for all theoretical results?
    Not Applicable
  \item Did you discuss competing hypotheses or theories that might challenge or complement your theoretical results?
    Not Applicable
  \item Have you considered alternative mechanisms or explanations that might account for the same outcomes observed in your study?
    Not Applicable
  \item Did you address potential biases or limitations in your theoretical framework?
    Not Applicable
  \item Have you related your theoretical results to the existing literature in social science?
    Not Applicable
  \item Did you discuss the implications of your theoretical results for policy, practice, or further research in the social science domain?
    Not Applicable
\end{enumerate}
We did not work with hypothesis testing. 

\item Additionally, if you are including theoretical proofs...
\begin{enumerate}
  \item Did you state the full set of assumptions of all theoretical results?
    Not Applicable
	\item Did you include complete proofs of all theoretical results?
    Not Applicable
\end{enumerate}

\item Additionally, if you ran machine learning experiments...
\begin{enumerate}
  \item Did you include the code, data, and instructions needed to reproduce the main experimental results (either in the supplemental material or as a URL)?
    Yes
  \item Did you specify all the training details (e.g., data splits, hyperparameters, how they were chosen)?
    Yes, both in Selected Models section and in the Appendix.
     \item Did you report error bars (e.g., with respect to the random seed after running experiments multiple times)?
    Yes
	\item Did you include the total amount of compute and the type of resources used (e.g., type of GPUs, internal cluster, or cloud provider)?
    Yes
     \item Do you justify how the proposed evaluation is sufficient and appropriate to the claims made? 
    Yes
     \item Do you discuss what is ``the cost`` of misclassification and fault (in)tolerance?
    Not Applicable
  
\end{enumerate}

\item Additionally, if you are using existing assets (e.g., code, data, models) or curating/releasing new assets, \textbf{without compromising anonymity}...
\begin{enumerate}
  \item If your work uses existing assets, did you cite the creators?
    Yes
  \item Did you mention the license of the assets?
    The datasets are all freely available and we reported their URLs. Same for the models employed in the experiments.
  \item Did you include any new assets in the supplemental material or as a URL?
    Not Applicable
  \item Did you discuss whether and how consent was obtained from people whose data you're using/curating?
    All the data are publicly available from their repositories.
  \item Did you discuss whether the data you are using/curating contains personally identifiable information or offensive content?
    Yes
\item If you are curating or releasing new datasets, did you discuss how you intend to make your datasets FAIR (see \cite{fair})?
Not Applicable
\item If you are curating or releasing new datasets, did you create a Datasheet for the Dataset (see \cite{gebru2021datasheets})? 
Not Applicable
\end{enumerate}

\item Additionally, if you used crowdsourcing or conducted research with human subjects, \textbf{without compromising anonymity}...
\begin{enumerate}
  \item Did you include the full text of instructions given to participants and screenshots?
    Not Applicable
  \item Did you describe any potential participant risks, with mentions of Institutional Review Board (IRB) approvals?
    Not Applicable
  \item Did you include the estimated hourly wage paid to participants and the total amount spent on participant compensation?
    Not Applicable
   \item Did you discuss how data is stored, shared, and deidentified?
   Not Applicable
\end{enumerate}

\end{enumerate}

\section{Appendix}\label{appendix}
\subsection*{Datasets URLs}\label{app:URLs}
In the following paragraph we list the datasets' URLs. 
\cite{Kiesel2019SemEval2019T4}: \url{https://zenodo.org/records/1489920}; the VIStA-H dataset 
 \cite{lyu_computational_2023}: \url{https://github.com/VIStA-H/Hyperpartisan-News-Titles/blob/main}; the Spanish Fake News Corpus \cite{gomez2021overview}: \url{https://github.com/jpposadas/FakeNewsCorpusSpanish}; the Fake News Net dataset \cite{Shu_FNNet}: \url{https://github.com/KaiDMML/FakeNewsNet/tree/master/dataset}; the Fake.br Corpus \cite{fakebr}: \url{https://github.com/roneysco/Fake.br-Corpus/blob/master/preprocessed}; CLEF 2022 CheckThat! Lab Subtask 1C \cite{CheckThat2022}: \url{https://gitlab.com/checkthat_lab/clef2022-checkthat-lab}; 
Qbias dataset \cite{Qbias} \url{https://github.com/irgroup/Qbias}; CLEF 2023 CheckThat! Lab Task 3A dataset \cite{azizovOverviewCLEF2023CheckThat} \url{https://gitlab.com/checkthat_lab/clef2023-checkthat-lab}.

\subsection{Computational Infrastructure and Hyperparameters}\label{sec:CIH}

Our computing infrastructure included two Tesla P40 GPUs, one NVIDIA GeForce RTX 2080 Ti GPU and a single A100 80GB SXM GPU, which was part of the Austral supercomputer of the CRIANN (Centre Régional Informatique et d’Applications Numériques de Normandie). Each experiment was run on a single GPU. 

\begin{table}[ht]
\centering
\renewcommand{\arraystretch}{1.2}
\begin{tabular}{|>{\raggedright\arraybackslash}m{4cm}|>{\centering\arraybackslash}m{4cm}|}
\hline
\textbf{Hyperparameter}       & \textbf{Value}       \\ \hline
Learning rate                 & $1 \times 10^{-4}$   \\ \hline
Epochs                        & 3                    \\ \hline
Runs                          & 5                    \\ \hline
LoRA target modules           & query, value         \\ \hline
LoRA Rank                     & 8                    \\ \hline
LoRA Alpha                    & 16                   \\ \hline
LoRA dropout                  & 0.1                  \\ \hline
Weight decay                  & 0.001                \\ \hline
Max grad norm                 & 0.3                  \\ \hline
Warmup ratio                  & 0.1                  \\ \hline
\end{tabular}
\caption{Hyperparameters for Fine-Tuning experiments.}
\label{tab:few shot_hyperparameters}
\end{table}

\subsection{Prompts}\label{app:prompts}
We employed four distinct prompt configurations in our experiments: (1) zero-shot prompts with a generic task definition, relying on the model’s internal knowledge; (2) zero-shot prompts with a task definition crafted by a political science expert; (3) task-specific prompts using structured codebooks; and (4) task-specific Chain-of-Thought (CoT) prompts.

For all configurations except the codebook-based prompts, we conducted preliminary experiments using the same structure but varying the wording and the label formatting. Based on these findings, we revised the label templates for the current study—most notably by switching from integer-based labels to string-based labels and adopting a more structured output format using special symbols (e.g., ==\textgreater) accompanied by a clear instruction to follow the format.

This refinement led to a substantial improvement in output consistency, reducing the number of unparseable labels by approximately 99\%.

\subsection{Prompt examples}\label{sec:new_prompts}

%PROMPT EXAMPLEs

%PROMPT HP

\onecolumn % This command switches the layout to a single column

\begin{longtable}{|>{\small}p{1.5cm}|>{\small}p{14.5cm}|}
\caption{Table showing different examples of instruction for PL Detection.}
\label{tab:PL_prompt_examples}\\
\hline
\textbf{Prompt} & \textbf{Text} \\
\hline
\endfirsthead

% This is the header that repeats on subsequent pages
\multicolumn{2}{|l|}{\tablename\ \thetable{} -- continued from previous page} \\
\hline
\textbf{Prompt} & \textbf{Text} \\
\hline
\endhead

% Your table rows go here:
zero shot generic& "You are an expert in political domain. You will be provided with a text, and your task is to determine, to the best of your judgment, whether the text is left-, right-wing or neutral. If you are unsure, make the most informed decision you can based on the content. Determine the political leaning of the headline. The possible choices are: 'left' if the article is left-wing, 'center' if the political leaning is from the center, 'right' if the article is right-wing'. Make sure to structure your answer in the following manner:\verb|\n\n|Final Answer ==\textgreater left\verb|\n\n|or \verb|\n\n|Final Answer ==\textgreater center \verb|\n\n|Final Answer ==\textgreater right\verb|\n\n| Here's the text: {}"
\\ \hline
zero shot generic& "You are an expert in political domain. You will be provided with a text, and your task is to determine, to the best of your judgment, whether the text is left-, right-wing or neutral. If you are unsure, make the most informed decision you can based on the content. Determine the political leaning of the headline. The possible choices are: 'left' if the article is left-wing, 'center' if the political leaning is from the center, 'right' if the article is right-wing'. Left-wing ideologies and moral values often include support for social equality, environmentalism, social justice, labor rights, and progressive taxation. Center ideologies and moral values often focus on a balance between market and social justice, pragmatism, and incremental reform. Right-wing ideologies and moral values often emphasize tradition, free market capitalism, individualism, and a limited role of government. Detect the political leaning of the headline. Determine the political leaning of the headline. The possible choices are: 'left' if the article is left-wing, 'center' if the political leaning is from the center, 'right' if the article is right-wing'. Make sure to structure your answer in the following manner:\verb|\n\n|Final Answer ==\textgreater left\verb|\n\n|or \verb|\n\n|Final Answer ==\textgreater center \verb|\n\n|Final Answer ==\textgreater right. \verb|\n\n| Here's the text: {}",
\\ \hline
few shot& "You are an expert in political domain. You will be provided with a text and some labeled examples, and your task is to determine, to the best of your judgment, whether the text is left-, right-wing or neutral. If you are unsure, make the most informed decision you can based on the content. Determine the political leaning of the headline. The possible choices are: 'left' if the article is left-wing, 'center' if the political leaning is from the center, 'right' if the article is right-wing'. Make sure to structure your answer in the following manner:\verb|\n\n|Final Answer ==\textgreater left\verb|\n\n|or \verb|\n\n|Final Answer ==\textgreater center \verb|\n\n|Final Answer ==\textgreater right\verb|\n\n| Here's the text: {}",
\\ \hline
codebook& "You are an expert in political domain. You will be provided with a codebook to follow and a text, and your task is to determine, to the best of your judgment, whether the text is left-, right-wing or neutral. If you are unsure, make the most informed decision you can based on the content. Determine the political leaning of the headline. \**Definition: Left-wing ideologies and moral values often include support for social equality, environmentalism, social justice, labor rights, and progressive taxation. Center ideologies and moral values often focus on a balance between market and social justice, pragmatism, and incremental reform. Right-wing ideologies and moral values often emphasize tradition, free market capitalism, individualism, and a limited role of government. \**Task: The possible choices are: 'left' if the article is left-wing, 'center' if the political leaning is from the center, 'right' if the article is right-wing'. Make sure to structure your answer in the following manner:\verb|\n\n|Final Answer ==\textgreater [left or center or right] \verb|\n\n| Here's the text: {}",
\\ \hline
cot & "You are an expert in political domain. You will be provided with a text, and your task is to determine, to the best of your judgment, whether the text is left-, right-wing or neutral. If you are unsure, make the most informed decision you can based on the content. Determine the political leaning of the headline. The possible choices are: 'left' if the article is left-wing, 'center' if the political leaning is from the center, 'right' if the article is right-wing'. Make sure to structure your answer in the following manner:\verb|\n\n|Final Answer ==\textgreater [left or center or right] Reason step by step: 1. Identify key words and phrases: list of words or phrases that might indicate political bias. 2. Sentiment analysis: Analyze the tone and the language to see if there are polarizing words, or emotional language. 3. Are there any rhetorical devices used? (e.g., hyperbole, loaded questions). 4. Look for policy implications: Does the headline suggest support for or opposition to specific policies? Are there any implicit assumptions about political or economic systems? 5. Identify target audience: Who seems to be the intended audience for this headline? 6. Left-wing hyperpartisan: Consider you are a left-wing reader. Would you consider this article as hyperpartisan to your political stance? [Follow the instructions from 1 to 5.] 7. Right-wing hyperpartisan: Consider you are a right-wing reader. Would you consider this article as hyperpartisan to your political stance? [Follow the instructions from 1 to 5.] 8. Consider omissions and emphasis: What information is emphasized, and what might be omitted? How might this emphasis or omission indicate bias? 9. Final prediction. The output will be in this format. Remember to use all reasoning steps followed by \textless end\textgreater to write a coincise explanation: Example output: ** Key words: EXPLANATION \textless end\textgreater ** Sentiment analysis: EXPLANATION \textless end\textgreater ** Policy implications: EXPLANATION \textless end\textgreater \** Target audience: EXPLANATION \textless end\textgreater ** Left-wing hyperpartisan: EXPLANATION \textless end\textgreater \** Right-wing hyperpartisan: EXPLANATION \textless end\textgreater ** Omissions and emphasis: EXPLANATION \textless end\textgreater \**Final Answer ==\textgreater left\verb|\n\n|or \verb|\n\n|Final Answer ==\textgreater center \verb|\n\n|Final Answer ==\textgreater right\verb|\n\n| Here's the text: {}"
\\ \hline
\end{longtable}

\onecolumn % This command switches the layout to a single column
\begin{longtable}{|>{\small}p{1.5cm}|>{\small}p{14.5cm}|}
\caption{Table showing different examples of instruction for HF Detection.}
\label{tab:PL_prompt_examples}\\
\hline
\textbf{Prompt} & \textbf{Text} \\
\hline
\endfirsthead

% This is the header that repeats on subsequent pages
\multicolumn{2}{|l|}{\tablename\ \thetable{} -- continued from previous page} \\
\hline
\textbf{Prompt} & \textbf{Text} \\
\hline
\endhead

% This is the footer that appears on the very last page

      zero shot generic& "You are a fake news expert. You will be provided with a text, and your task is to determine, to the best of your judgment, whether the text is harmful to society or not. If you are unsure, make the most informed decision you can based on the content. If the news is harmful, the label is 'harmful', if the text is neutral, the label is 'neutral'. Make sure to structure your answer in the following manner:\verb|\n\n| Final Answer ==\textgreater harmful\verb|\n\n|or \verb|\n\n|Final Answer ==\textgreater neutral\verb|\n\n| Here's the text: {}",

\\ \hline
      zero shot generic& "You are a fake news expert. Harmful tweet is a post on the social media platform Twitter that can cause distress, harm, or damage to individuals, groups, or society. This type of tweet may include content that is abusive, offensive, defamatory, threatening, or incites violence against the policies to contain the spread of COVID-19. Additionally, it may spread false information, contribute to cyberbullying, or perpetuate hate speech and discrimination. Harmful tweets can negatively impact mental health, reputations, and community safety. You will be provided with labeled examples and a text, and your task is to determine, to the best of your judgment, whether the text is harmful to society or not. If you are unsure, make the most informed decision you can based on the content. If the news is harmful, the label is 'harmful', if the text is a neutral, the label is 'neutral'. Make sure to structure your answer in the following manner:\verb|\n\n| Final Answer ==\textgreater harmful\verb|\n\n|or\verb|\n\n|Final Answer ==\textgreater neutral\verb|\n\n| Here's the text: {}",

\\ \hline
      few shot& "You are a fake news expert. You will be provided with labeled examples and a text, and your task is to determine, to the best of your judgment, whether the text is harmful to society or not. If you are unsure, make the most informed decision you can based on the content. If the news is harmful, the label is 'harmful', if the text is neutral, the label is 'neutral'. Make sure to structure your answer in the following manner:\verb|\n\n| Final Answer ==\textgreater harmful\verb|\n\n|or\verb|\n\n|Final Answer ==\textgreater neutral\verb|\n\n|Here's the text: {}",

 \\ \hline
      codebook& "You are a fake news expert. **Definition: Harmful tweet is a post on the social media platform Twitter that can cause distress, harm, or damage to individuals, groups, or society. This type of tweet may include content that is abusive, offensive, defamatory, threatening, or incites violence against the policies to contain the spread of COVID-19. Additionally, it may spread false information, contribute to cyberbullying, or perpetuate hate speech and discrimination. Harmful tweets can negatively impact mental health, reputations, and community safety. **Task: You will be provided with a codebook and a text, and your task is to determine, to the best of your judgment, whether the text is harmful to society or not. If you are unsure, make the most informed decision you can based on the content. If the news is harmful, the label is 'harmful', if the text is a neutral, the label is 'neutral'. Make sure to structure your answer in the following manner:\verb|\n\n| Final Answer ==\textgreater harmful\verb|\n\n|or \verb|\n\n|Final Answer ==\textgreater neutral\verb|\n\n| Here's the text: {} \verb|\n\n|This is the 
      codebook: | Label | Class | Definition | Typical Content Patterns | Neutral | Neutral| The tweet provides factual, supportive, or neutral information about COVID-19. | - Promotes vaccinations, masking, or safety |\verb|\n||- Shares news from reliable sources (WHO, CDC) |\verb|\n||- Describes personal experiences without disinformation |\verb|\n|| Harmful | Harmful | The tweet contains disinformation, conspiracy theories, or medically incorrect claims. | - Claims vaccines are deadly or part of a conspiracy |\verb|\n||- Encourages rejection of public health guidance |\verb|\n||- Spreads pseudoscience or political hoaxes |\verb|\n\n|| Aspect | Description of Harmful Characteristics  Conspiracy Theory | References to population control, “pandemic”, 5G causes COVID, microchips in vaccines |\verb|\n|| Distrust in Science | Discredits CDC/WHO, claims that mainstream medicine is false or corrupt |\verb|\n|| Alternative Cures | Promotes ivermectin, bleach, natural remedies as vaccine alternatives without evidence |\verb|\n|| Anti-Vaccine | Claims vaccines are unsafe, rushed, or part of a control agenda |\verb|\n|| Hoax Narrative | Declares COVID-19 is fake, exaggerated, or invented by governments |\verb|\n|| Call to Action | Urges followers to ignore mandates, avoid vaccination, or resist masking |\verb|\n|| Language Style | Emotionally charged, sarcastic, or mocking tone; often uses ALL CAPS, emojis, hashtags like **NoVax **Plandemic |\verb|\n\n|| Linguistic Features | Harmful | Not Harmful | Use of hashtags | **plandemic, **NoVax, **scamdemic | **GetVaccinated, **StaySafe ||\verb|\n||| Source credibility | Anonymous sources, influencer claims | WHO, CDC, news outlets ||\verb|\n|| Personal stories | Claims of side effects or hoaxes | Stories of recovery, gratitude for vaccines ||\verb|\n|| Tone | Alarmist, rebellious, sarcastic | Informative, cautious, or appreciative |. \verb|\n\n|Make sure to structure your answer in the following manner:\verb|\n\n| Final Answer ==\textgreater harmful\verb|\n\n|or\verb|\n\n|Final Answer ==\textgreater neutral\verb|\n\n| Here's the text: {}",
 \\ \hline
  cot& "You are a fake news expert. You will be provided with labeled examples and a text, and your task is to determine, to the best of your judgment, whether the text is harmful to society or not. If you are unsure, make the most informed decision you can based on the content. Analyze the text for fake news using step-by-step reasoning. If the news is neutral, the label is 'neutral', if the text is harmful, the label is 'harmful'. You have to choose one label and the output is only one label. Make sure to structure your answer in the following manner: Explanation: [Eplanation for each reasoning step]. \verb|\n\n| Final Answer ==\textgreater harmful |\verb|\n|| or |\verb|\n|| Final Answer ==\textgreater eutral.|\verb|\n||****** Reason step by step:|\verb|\n||1. *Lexical Analysis**\verb|\n\n|- Check for vague sourcing, absolutist language, logical fallacies|\verb|\n||. **Sentiment Analysis**\verb|\n\n|- Identify polarizing/emotional language|\verb|\n||3. **Content Authenticity**\verb|\n\n|- Detect bot-like patterns, verify data|\verb|\n||4. **Framing Bias**\verb|\n\n|- Assess selective presentation|\verb|\n|| 5. Audience Analysis**\verb|\n\n|- Determine target audience and intent|\verb|\n||6. **Information Adequacy**\verb|\n\n|- Evaluate comprehensiveness and reliability. \verb|\n\n| Make sure to structure your answer in the following manner:Explanation: [Explanation for each reasoning step]. Final Answer ==\textgreater harmful\verb|\n\n|or\verb|\n\n|Final Answer ==\textgreater neutral. 
  \verb|\n\n| Here's the text: {}"

 \\ \hline
 \caption{Table showing different examples of instruction for HF Detection. For the multilingual dataset, we translated this prompt to Arabic and Bulgarian.}
\label{tab:HF_prompt_examples}
\end{longtable}

\onecolumn % This command switches the layout to a single column

\begin{longtable}{|>{\small}p{1.5cm}|>{\small}p{14.5cm}|}
\caption{Table showing different examples of instruction for HP Detection.}
\label{tab:PL_prompt_examples}\\
\hline
\textbf{Prompt} & \textbf{Text} \\
\hline
\endfirsthead

% This is the header that repeats on subsequent pages
\multicolumn{2}{|l|}{\tablename\ \thetable{} -- continued from previous page} \\
\hline
\textbf{Prompt} & \textbf{Text} \\
\hline
\endhead
zero shot generic &      "You are a hyperpartisan news expert. You will be provided with a text, and your task is to determine, to the best of your judgment, whether the text is hyperpartisan or not. If you are unsure, make the most informed decision you can based on the content. The possible choices are: 'neutral' if the article is neutral, 'hyperpartisan' if the article is hyperpartisan. You have to choose one label and the output is only one label. \verb|\n\n| Make sure to structure your answer in the following manner:\verb|\n\n| Final Answer ==\textgreater hyperpartisan\verb|\n\n|or \verb|\n\n|Final Answer ==\textgreater neutral \verb|\n\n| This is the text: {}" \\ \hline
"zero shot generic"& "You are a hyperpartisan news expert. You will be provided with a text, and your task is to determine, to the best of your judgment, whether the text is hyperpartisan or not. If you are unsure, make the most informed decision you can based on the content. Hyperpartisan articles contain biases, particularly ad hominem attack, loaded language, and evidence of political ideology. Sometimes they rely on cherry-picking strategy. The possible choices are: 'neutral' if the article is neutral, 'hyperpartisan' if the article is hyperpartisan. You have to choose one label and the output is only one label. Make sure to structure your answer in the following manner:\verb|\n\n| Final Answer ==\textgreater hyperpartisan\verb|\n\n|or \verb|\n\n|Final Answer ==\textgreater neutral \verb|\n\n| This is the text: {}" \\ \hline
few shot & "You are a hyperpartisan news expert. You will be provided with a text, and your task is to determine, to the best of your judgment, whether the text is hyperpartisan or not. If you are unsure, make the most informed decision you can based on the content. You will be provided with some examples of labeled text. The possible choices are: 'neutral' if the article is neutral, 'hyperpartisan' if the article is hyperpartisan. You have to choose one label and the output is only one label. Make sure to structure your answer in the following manner:\verb|\n\n| Final Answer ==\textgreater hyperpartisan\verb|\n\n|or \verb|\n\n|Final Answer ==\textgreater neutral \verb|\n\n|******Examples of labelled articles: {} \verb|\n\n| This is the text: {}" \\ \hline

codebook& "You are a hyperpartisan news expert. You will be provided with a text, and your task is to determine, to the best of your judgment, whether the text is hyperpartisan or not. If you are unsure, make the most informed decision you can based on the content. You will be provided with some examples of labeled text. \verb|\n\n| Definition: Hyperpartisan news detection is the process of identifying news articles that exhibit extreme one-sidedness, characterized by a pronounced use of bias. The prefix hyper- highlights the exaggerated application of at least one specific type of bias—such as spin, ad hominem attacks, opinionated statements, ideological slants, framing, selective coverage, political leaning, or slant bias—to promote a particular ideological perspective. This strong ideological alignment is conveyed through amplified linguistic elements that reinforce one of these bias types within the text.\verb|\n\n|****** Task: Read carefully the codebook provided and assign a label to the text. You can choose only one label. If the text is neutral, you will write 'neutral', if it is hyperpartisan 'hyperpartisan'.\verb|\n\n|Follow the output template given as an example. Under no circumstances we are asking to provide or generate harmful content. Please, provide only the label.\verb|\n\n|****** Make sure to structure your answer in the following manner:\verb|\n\n| Final Answer ==\textgreater hyperpartisan\verb|\n\n|or \verb|\n\n|Final Answer ==\textgreater neutral \verb|\n\n| **This is the codebook: Linguistic Features:\verb|\n\n|******** 1. Lexical Features |\verb|\n| Feature | Hyperpartisan | Neutral | Lexical Polarity | Frequent use of emotionally charged words (e.g., disaster, outrageous) | Neutral and precise language ||\verb|\n||| Modality and Certainty | Strong modal verbs (e.g., will destroy) | Hedging markers (e.g., may, might) ||\verb|\n||| Repetition | Repeating claims or slogans | Minimal repetition ||\verb|\n||| Pronouns | Frequent us vs them language | Focus on third-person objectivity |\verb|\n\n|******** 2. Rhetorical Devices|\verb|\n||| Feature | Hyperpartisan | Neutral ||\verb|\n||| Appeal to Emotion | Frequent appeals to fear/anger | Logical/factual appeal ||\verb|\n||| Hyperbole | Common exaggeration | Proportional statements ||\verb|\n||| Metaphors | Politically loaded metaphors | Literal language preferred |\verb|\n\n|******** 3. Discourse Structure|\verb|\n||| Feature | Hyperpartisan | Neutral ||\verb|\n||| Framing | Blame/conflict framing | Balanced framing |\verb|\n|| Source Attribution | Partisan sources only | Multiple reputable sources ||\verb|\n||| Balance of Views | One-sided presentation | Multiple perspectives |\verb|\n\n|******** 4. Ideological Markers |\verb|\n||| Feature | Hyperpartisan | Neutral |\verb|\n|| Us vs Them | Strong binary division | Avoids binaries |\verb|\n|| Ideological Alignment | Clear left/right alignment | Issue-focused |\verb|\n\n|******** 5. Pragmatic Features|\verb|\n||| Feature | Hyperpartisan | Neutral ||\verb|\n||| Intent | Persuade/convert | Inform/explain ||\verb|\n||| Tone | Confrontational/accusatory | Formal/detached |\verb|\n\n|****** \verb|\n\n| Final Answer ==\textgreater hyperpartisan\verb|\n\n|or \verb|\n\n|Final Answer ==\textgreater neutral\verb|\n\n| This is the text: {}" \\ \hline

cot& "You are a hyperpartisan news expert. You will be provided with a text, and your task is to determine, to the best of your judgment, whether the text is hyperpartisan or not. If you are unsure, make the most informed decision you can based on the content. You have to choose one label and the output will be the explanation and the determined label for that article. Make sure to structure your answer in the following manner:\verb|\n\n| Explanation: [Eplanation for each reasoning step]. \verb|\n\n| Final Answer ==\textgreater hyperpartisan\verb|\n\n|or \verb|\n\n|Final Answer ==\textgreater neutral. \verb|\n\n| Reason step by step:\verb|\n\n|1. **Lexical and Sentiment analysis**: Analyze the tone and language. Does the article use polarizing, emotionally charged, or exaggerated language? Are there strong positive/negative sentiments toward a group, ideology, or issue?\verb|\n\n|2. **Rhetorical bias**: Does the text use loaded language, name-calling, or manipulative rhetoric? Are there oversimplifications, strawman arguments, or exaggerated claims?\verb|\n\n|3. **Framing bias**: Is information presented selectively to favor one perspective? Does it emphasize certain aspects while downplaying others to shape perception?\verb|\n\n|4. **Ideological bias**: Does it emphasize certain aspects while downplaying others to shape perception? Does the article vilify opposing views rather than engaging with them fairly?\verb|\n\n|5. **Unilateral coverage**: Does the article present multiple viewpoints (neutral) or only one side (hyperpartisan)? Are opposing arguments ignored, misrepresented, or dismissed?\verb|\n\n|6. **Intent and Purpose**: Is the primary goal to inform objectively (neutral) or to persuade/mislead (hyperpartisan)? Does it present facts fairly, or does it push a clear agenda?\verb|\n\n|7. Final prediction: Based on your previous considerations, classify the input as:\verb|\n\n|- 'neutral' (Neutral): Balanced, factual, and objective.\verb|\n\n|- 'hyperpartisan': Biased, one-sided, or manipulative.\verb|\n\n| While generating the explanation for each reasoning step be coincise.. \verb|\n\n| Remember to follow the output template for the label. \verb|\n\n| Final Answer ==\textgreater hyperpartisan\verb|\n\n|or \verb|\n\n|Final Answer ==\textgreater neutral\verb|\n\n| This is the text: {}" \\ \hline
\caption{Table showing different examples of instruction for HP Detection.}
\end{longtable}

\label{tab:HP_prompt_examples}

\onecolumn % This command switches the layout to a single column

\begin{longtable}{|>{\small}p{1.5cm}|>{\small}p{14.5cm}|}
\caption{Table showing different examples of instruction for FN Detection.}
\label{tab:PL_prompt_examples}\\
\hline
\textbf{Prompt} & \textbf{Text} \\
\hline
\endfirsthead

% This is the header that repeats on subsequent pages
\multicolumn{2}{|l|}{\tablename\ \thetable{} -- continued from previous page} \\
\hline
\textbf{Prompt} & \textbf{Text} \\
\hline
\endhead
      zero shot generic & "You are a fake news expert. You will be provided with a text, and your task is to determine, to the best of your judgment, whether the text is fake news or not. If you are unsure, make the most informed decision you can based on the content. If the news is true, the label is 'true', if the text is a fake news, the label is 'fake'. Make sure to structure your answer in the following manner:\verb|\n\n| Final Answer ==\textgreater fake\verb|\n\n|or \verb|\n\n|Final Answer ==\textgreater true\verb|\n\n| Here's the text: {}"
\\ \hline
      zero shot generic& "You are a fake news expert. We define fake news as: \"purposefully crafted, sensational, emotionally charged, misleading or totally fabricated information that mimics mainstream news\". You will be provided with a text and your task is to determine, to the best of your judgment, whether the text is fake news or not. If you are unsure, make the most informed decision you can based on the content. If the news is true, the label is 'true', if the text is a fake news, the label is 'fake'. Make sure to structure your answer in the following manner:\verb|\n\n| Final Answer ==\textgreater fake\verb|\n\n|or \verb|\n\n|Final Answer ==\textgreater true\verb|\n\n| Here's the text: {}",
\\ \hline
      few shot& "You are a fake news expert. You will be provided with a text and labeled examples, and your task is to determine, to the best of your judgment, whether the text is fake news or not. If you are unsure, make the most informed decision you can based on the content. If the news is true, the label is 'true', if the text is a fake news, the label is 'fake'. Make sure to structure your answer in the following manner:\verb|\n\n| Final Answer ==\textgreater fake\verb|\n\n|or \verb|\n\n|Final Answer ==\textgreater true\verb|\n\n| Here's the text: {}",
 \\ \hline
  codebook& "You are a fake news expert. You will be provided with a text, and your task is to determine, to the best of your judgment, whether the text is fake news or not. If you are unsure, make the most informed decision you can based on the content. **Definition: Fake news detection identifies intentionally misleading content characterized by sensationalism, lack of credible sources, or manipulative language.\verb|\n\n|If the news is true, the label is 'true', if the text is a fake news, the label is 'fake'. \verb|\n\n|****** Task: To structure the output, follow the template in the example. \verb|\n\n| Make sure to structure your answer in the following manner:\verb|\n\n| Final Answer ==\textgreater fake\verb|\n\n|or \verb|\n\n|Final Answer ==\textgreater true. This is the codebook you have to use to perform the classification: \verb|\n\n|****** Detection Criteria:\verb|\n|1. **Source Origin**\verb|\n\n|- Real: Established media, government sources\verb|\n\n|- Fake: Anonymous/unverifiable sources\verb|\n|2. **Event Reporting**\verb|\n\n|- Real: Specific references, quantitative data\verb|\n\n|- Fake: Broad generalizations, unverifiable claims\verb|\n|3. **Language Style**\verb|\n\n|- Real: Neutral,   professional\verb|\n\n|- Fake: Sensationalized, lickbait\verb|\n|4. **Entity Authenticity**\verb|\n\n|- Real: Named real-world entities\verb|\n\n|- Fake: Fictional/misspelled names\verb|\n|5. **Claim Reliability**\verb|\n\n|- Real: Evidence-based\verb|\n\n|- Fake: Absurd/absolute claims\verb|\n|6.   **Emotional Tone**\verb|\n\n|- Real: Objective\verb|\n\n|- Fake: Emotional intensifiers\verb|\n|7. **Source Credibility**\verb|\n\n|- Real: Verifiable\verb|\n\n|- Fake: Unknown/misleading domains \verb|\n|8. **Political   Balance**\verb|\n\n|- Real: Multi-perspective\verb|\n\n|- Fake: One-sided\verb|\n|9. **Satire Markers**\verb|\n\n|- Real: No  satire \verb|\n|- Fake: Absurd content\verb|\n|10. *Conspiracy Indicators* \verb|\n|- Real: Supported theories\verb|\n\n|- Fake: Fringe conspiracy phrases\verb|\n\n| Make sure to structure your answer in the following manner:\verb|\n\n| Final Answer ==\textgreater fake\verb|\n\n|or \verb|\n\n|Final Answer ==\textgreater true \verb|\n\n| Here's the text: {}",
 \\ \hline
      cot& "You are a fake news expert. You will be provided with a text, and your task is to determine, to the best of your judgment, whether the text is fake news or not. If you are unsure, make the most informed decision you can based on the content. Analyze the text for fake news using step-by-step reasoning. If the news is true, the label is 'true', if the text is a fake news, the label is 'fake'. You have to choose one label and the output is only one label. To structure the output, follow the template in the example. \verb|\n\n|Output example: Explanation: [Eplanation for each reasoning step]. Final Prediction: Final Answer ==\textgreater [true or fake]. \verb|\n\n|****** Reasoning step by step:\verb|\n|1. **Lexical Analysis**\verb|\n\n|- Check for vague sourcing, absolutist language, logical fallacies\verb|\n|2. **Sentiment Analysis**\verb|\n\n|- Identify polarizing/emotional language\verb|\n|3. **Content Authenticity**\verb|\n\n|- Detect bot-like patterns, verify data\verb|\n|4. **Framing Bias**\verb|\n\n|- Assess selective presentation\verb|\n|5. 
      Audience Analysis**\verb|\n\n|- Determine target audience and intent\verb|\n|6. **Information Adequacy**\verb|\n\n|- Evaluate comprehensiveness and reliability. \verb|\n\n| Make sure to structure your answer in the following manner:\verb|\n\n| Final Answer ==\textgreater fake\verb|\n\n|or \verb|\n\n|Final Answer ==\textgreater true \verb|\n\n| Here's the text: {}" \\ \hline
      \caption{Table showing different examples of instruction for FN Detection. For the multilingual dataset, we translated this prompt to Spanish and Portugues.}
\end{longtable}

\label{tab:FN_prompt_examples}

\subsection*{Results and SOTA}\label{app:results_SOTA}

\begin{table}[htbp]
\rotatebox{90}{%
  \begin{minipage}{\textheight}
\centering
\resizebox{\textheight}{!}{%
\begin{tabular}{p{1cm}lrrrr||rrrr||rrrrrr||rrrrrr}
\toprule
 &  & \multicolumn{4}{c}{\textbf{Hyperpartisan Det.}} & \multicolumn{4}{c}{\textbf{Political Leaning Det.}} & \multicolumn{6}{c}{\textbf{Harmful Text Det.}} & \multicolumn{6}{c}{\textbf{Fake News Det.}} \\
 &  & \multicolumn{2}{c}{\textbf{SH}} & \multicolumn{2}{c}{\textbf{HV}} & \multicolumn{2}{c}{\textbf{C3A}} & \multicolumn{2}{c}{\textbf{QB}} & \multicolumn{2}{c}{\textbf{C1E}} & \multicolumn{2}{c}{\textbf{C1A}} & \multicolumn{2}{c}{\textbf{C1B}} & \multicolumn{2}{c}{\textbf{FNN}} & \multicolumn{2}{c}{\textbf{FBC}} & \multicolumn{2}{c}{\textbf{SFN}} \\
 &  & F1 & Acc & F1 & Acc & F1 & Acc & F1 & Acc & F1 & Acc & F1 & Acc & F1 & Acc & F1 & Acc & F1 & Acc & F1 & Acc \\
\cmidrule(lr){2-22}
\multirow{12}{*}{\parbox[c][12\baselineskip][c]{1cm}{\centering\rotatebox{90}{\textbf{Zero-Shot}}}} & \textbf{Generic} &  &  &  &  &  &  &  &  &  &  &  &  &  &  &  &  &  &  &  &  \\
 &   Llama-3.1-8B-Instruct & 0.678 & 0.701 & 0.742 & 0.739 & \underline{0.437} & \underline{0.425} & \underline{0.426} & \underline{0.425} & 0.730 & 0.689 & 0.722 & 0.679 & 0.507 & 0.372 & 0.364 & 0.329 & 0.240 & 0.256 & 0.208 & 0.210 \\
 &   Mistral-Nemo-Instruct-2407 & 0.686 & 0.707 & 0.607 & 0.623 & 0.422 & 0.421 & 0.360 & 0.360 & 0.720 & 0.677 & 0.565 & 0.516 & 0.793 & 0.695 & 0.551 & 0.542 & 0.304 & 0.327 & 0.144 & 0.145 \\
 &   Qwen2.5-7B-Instruct & 0.780 & 0.782 & 0.621 & 0.678 & 0.404 & 0.399 & 0.336 & 0.335 & 0.783 & 0.761 & \underline{\textbf{0.851}} & \underline{\textbf{0.857}} & 0.755 & 0.643 & 0.328 & 0.321 & 0.263 & 0.265 & \underline{0.469} & \underline{0.335} \\
\cmidrule(lr){2-22}
 & \textbf{Specific} &  &  &  &  &  &  &  &  &  &  &  &  &  &  &  &  &  &  &  &  \\
 &   Llama-3.1-8B-Instruct & 0.738 & 0.748 & \underline{0.764} & \underline{0.764} & 0.431 & 0.419 & 0.419 & 0.417 & 0.764 & 0.737 & 0.721 & 0.678 & 0.676 & 0.548 & 0.412 & 0.371 & 0.269 & 0.326 & 0.182 & 0.182 \\
 &   Mistral-Nemo-Instruct-2407 & 0.740 & 0.750 & 0.574 & 0.593 & 0.409 & 0.408 & 0.352 & 0.356 & 0.719 & 0.677 & 0.613 & 0.562 & 0.818 & 0.732 & 0.549 & 0.540 & 0.315 & 0.354 & 0.150 & 0.150 \\
 &   Qwen2.5-7B-Instruct & 0.789 & 0.790 & 0.585 & 0.668 & 0.395 & 0.391 & 0.340 & 0.335 & 0.721 & 0.685 & 0.827 & 0.818 & 0.738 & 0.622 & 0.257 & 0.278 & 0.275 & 0.276 & 0.194 & 0.205 \\
\cmidrule(lr){2-22}
 & \textbf{Codebook} &  &  &  &  &  &  &  &  &  &  &  &  &  &  &  &  &  &  &  &  \\
 &   Llama-3.1-8B-Instruct & 0.748 & 0.753 & 0.715 & 0.722 & 0.382 & 0.407 & 0.324 & 0.344 & 0.762 & 0.729 & 0.728 & 0.686 & 0.670 & 0.538 & 0.388 & 0.349 & 0.304 & 0.405 & 0.165 & 0.166 \\
 &   Mistral-Nemo-Instruct-2407 & 0.694 & 0.713 & 0.706 & 0.704 & 0.372 & 0.385 & 0.318 & 0.340 & 0.753 & 0.717 & 0.636 & 0.585 & \underline{\textbf{0.864}} & \underline{\textbf{0.806}} & \underline{0.580} & \underline{0.595} & \underline{0.335} & \underline{\textbf{0.440}} & 0.144 & 0.145 \\
 &   Qwen2.5-7B-Instruct & \underline{\textbf{0.810}} & \underline{\textbf{0.811}} & 0.636 & 0.688 & 0.377 & 0.395 & 0.311 & 0.323 & \underline{0.785} & \underline{0.769} & 0.832 & 0.838 & 0.820 & 0.735 & 0.341 & 0.320 & 0.256 & 0.256 & 0.422 & 0.295 \\
\midrule
\midrule
\multirow{68}{*}{\parbox[c][68\baselineskip][c]{1cm}{\centering\rotatebox{90}{\textbf{Few-Shot}}}} & \textbf{Random 1-shot} &  &  &  &  &  &  &  &  &  &  &  &  &  &  &  &  &  &  &  &  \\
 &   Llama-3.1-8B-Instruct & 0.751 $\pm$0.037 & 0.755 $\pm$0.034 & 0.758 $\pm$0.015 & 0.756 $\pm$0.015 & 0.463 $\pm$0.022 & 0.457 $\pm$0.024 & 0.411 $\pm$0.003 & 0.411 $\pm$0.006 & 0.725 $\pm$0.026 & 0.684 $\pm$0.032 & 0.745 $\pm$0.036 & 0.708 $\pm$0.044 & 0.580 $\pm$0.047 & 0.446 $\pm$0.047 & 0.288 $\pm$0.033 & 0.270 $\pm$0.023 & 0.257 $\pm$0.028 & 0.305 $\pm$0.036 & 0.184 $\pm$0.006 & 0.185 $\pm$0.006 \\
 &   Mistral-Nemo-Instruct-2407 & 0.716 $\pm$0.017 & 0.728 $\pm$0.014 & 0.776 $\pm$0.021 & 0.774 $\pm$0.021 & 0.406 $\pm$0.011 & 0.390 $\pm$0.013 & 0.393 $\pm$0.014 & 0.395 $\pm$0.009 & 0.679 $\pm$0.024 & 0.629 $\pm$0.028 & 0.511 $\pm$0.110 & 0.473 $\pm$0.091 & 0.827 $\pm$0.067 & 0.752 $\pm$0.096 & \underline{0.570 $\pm$0.028} & \underline{0.580 $\pm$0.049} & 0.298 $\pm$0.011 & 0.334 $\pm$0.022 & 0.155 $\pm$0.005 & 0.156 $\pm$0.006 \\
 &   Qwen2.5-7B-Instruct & 0.718 $\pm$0.041 & 0.730 $\pm$0.035 & 0.615 $\pm$0.025 & 0.683 $\pm$0.013 & 0.394 $\pm$0.020 & 0.402 $\pm$0.014 & 0.356 $\pm$0.007 & 0.355 $\pm$0.006 & \underline{\textbf{0.834 $\pm$0.010}} & \underline{\textbf{0.838 $\pm$0.015}} & 0.820 $\pm$0.009 & 0.854 $\pm$0.005 & 0.810 $\pm$0.035 & 0.723 $\pm$0.051 & 0.248 $\pm$0.031 & 0.273 $\pm$0.011 & 0.306 $\pm$0.011 & 0.332 $\pm$0.021 & 0.629 $\pm$0.016 & 0.640 $\pm$0.012 \\
\cmidrule(lr){2-22}
 & \textbf{Random 2-shot} &  &  &  &  &  &  &  &  &  &  &  &  &  &  &  &  &  &  &  &  \\
 &   Llama-3.1-8B-Instruct & 0.664 $\pm$0.041 & 0.685 $\pm$0.031 & 0.767 $\pm$0.020 & 0.765 $\pm$0.019 & 0.457 $\pm$0.021 & 0.451 $\pm$0.018 & 0.412 $\pm$0.007 & 0.416 $\pm$0.008 & 0.746 $\pm$0.012 & 0.708 $\pm$0.015 & 0.712 $\pm$0.034 & 0.668 $\pm$0.039 & 0.422 $\pm$0.066 & 0.302 $\pm$0.052 & 0.300 $\pm$0.027 & 0.278 $\pm$0.020 & 0.241 $\pm$0.018 & 0.269 $\pm$0.019 & 0.179 $\pm$0.011 & 0.179 $\pm$0.011 \\
 &   Mistral-Nemo-Instruct-2407 & 0.725 $\pm$0.013 & 0.735 $\pm$0.009 & 0.641 $\pm$0.069 & 0.653 $\pm$0.057 & 0.421 $\pm$0.011 & 0.407 $\pm$0.011 & 0.404 $\pm$0.018 & 0.400 $\pm$0.017 & 0.626 $\pm$0.030 & 0.574 $\pm$0.032 & 0.619 $\pm$0.066 & 0.571 $\pm$0.066 & 0.797 $\pm$0.032 & 0.703 $\pm$0.045 & 0.523 $\pm$0.028 & 0.505 $\pm$0.041 & 0.276 $\pm$0.020 & 0.296 $\pm$0.018 & 0.153 $\pm$0.007 & 0.155 $\pm$0.007 \\
 &   Qwen2.5-7B-Instruct & 0.788 $\pm$0.018 & 0.789 $\pm$0.017 & 0.661 $\pm$0.026 & 0.711 $\pm$0.014 & 0.408 $\pm$0.012 & 0.411 $\pm$0.010 & 0.368 $\pm$0.009 & 0.361 $\pm$0.009 & 0.814 $\pm$0.011 & 0.804 $\pm$0.013 & 0.838 $\pm$0.007 & \underline{0.855 $\pm$0.003} & 0.758 $\pm$0.012 & 0.648 $\pm$0.016 & 0.252 $\pm$0.030 & 0.277 $\pm$0.013 & 0.282 $\pm$0.004 & 0.299 $\pm$0.013 & 0.619 $\pm$0.033 & 0.640 $\pm$0.022 \\
\cmidrule(lr){2-22}
 & \textbf{Random 3-shot} &  &  &  &  &  &  &  &  &  &  &  &  &  &  &  &  &  &  &  &  \\
 &   Llama-3.1-8B-Instruct & 0.654 $\pm$0.020 & 0.678 $\pm$0.017 & 0.780 $\pm$0.018 & 0.778 $\pm$0.019 & 0.466 $\pm$0.016 & 0.461 $\pm$0.019 & 0.434 $\pm$0.009 & 0.428 $\pm$0.011 & 0.732 $\pm$0.024 & 0.692 $\pm$0.029 & 0.701 $\pm$0.034 & 0.655 $\pm$0.039 & 0.544 $\pm$0.086 & 0.412 $\pm$0.078 & 0.285 $\pm$0.030 & 0.268 $\pm$0.021 & 0.270 $\pm$0.015 & 0.331 $\pm$0.028 & 0.179 $\pm$0.005 & 0.180 $\pm$0.005 \\
 &   Mistral-Nemo-Instruct-2407 & 0.733 $\pm$0.019 & 0.740 $\pm$0.016 & 0.678 $\pm$0.062 & 0.684 $\pm$0.056 & 0.415 $\pm$0.008 & 0.404 $\pm$0.009 & 0.375 $\pm$0.019 & 0.374 $\pm$0.015 & 0.633 $\pm$0.018 & 0.581 $\pm$0.019 & 0.613 $\pm$0.040 & 0.563 $\pm$0.039 & 0.789 $\pm$0.035 & 0.693 $\pm$0.050 & 0.523 $\pm$0.031 & 0.507 $\pm$0.043 & 0.303 $\pm$0.014 & 0.309 $\pm$0.017 & 0.156 $\pm$0.005 & 0.157 $\pm$0.005 \\
 &   Qwen2.5-7B-Instruct & 0.777 $\pm$0.014 & 0.780 $\pm$0.013 & 0.679 $\pm$0.010 & 0.720 $\pm$0.007 & 0.417 $\pm$0.012 & 0.421 $\pm$0.005 & 0.370 $\pm$0.006 & 0.359 $\pm$0.005 & 0.809 $\pm$0.013 & 0.799 $\pm$0.018 & 0.846 $\pm$0.007 & 0.852 $\pm$0.004 & 0.748 $\pm$0.025 & 0.634 $\pm$0.033 & 0.245 $\pm$0.022 & 0.271 $\pm$0.010 & 0.308 $\pm$0.009 & 0.323 $\pm$0.013 & 0.652 $\pm$0.040 & 0.663 $\pm$0.032 \\
\cmidrule(lr){2-22}
 & \textbf{Random 4-shot} &  &  &  &  &  &  &  &  &  &  &  &  &  &  &  &  &  &  &  &  \\
 &   Llama-3.1-8B-Instruct & 0.623 $\pm$0.037 & 0.656 $\pm$0.028 & 0.762 $\pm$0.035 & 0.760 $\pm$0.035 & 0.463 $\pm$0.020 & 0.456 $\pm$0.022 & 0.425 $\pm$0.010 & 0.421 $\pm$0.007 & 0.739 $\pm$0.020 & 0.700 $\pm$0.024 & 0.688 $\pm$0.024 & 0.640 $\pm$0.027 & 0.509 $\pm$0.090 & 0.380 $\pm$0.081 & 0.275 $\pm$0.035 & 0.264 $\pm$0.023 & 0.234 $\pm$0.018 & 0.267 $\pm$0.047 & 0.180 $\pm$0.012 & 0.183 $\pm$0.013 \\
 &   Mistral-Nemo-Instruct-2407 & 0.741 $\pm$0.025 & 0.747 $\pm$0.020 & 0.614 $\pm$0.064 & 0.631 $\pm$0.052 & 0.417 $\pm$0.005 & 0.404 $\pm$0.007 & 0.396 $\pm$0.017 & 0.392 $\pm$0.013 & 0.610 $\pm$0.017 & 0.558 $\pm$0.017 & 0.642 $\pm$0.043 & 0.593 $\pm$0.044 & 0.790 $\pm$0.033 & 0.694 $\pm$0.049 & 0.500 $\pm$0.043 & 0.475 $\pm$0.059 & 0.290 $\pm$0.014 & 0.291 $\pm$0.014 & 0.168 $\pm$0.003 & 0.172 $\pm$0.004 \\
 &   Qwen2.5-7B-Instruct & 0.779 $\pm$0.009 & 0.780 $\pm$0.009 & 0.681 $\pm$0.006 & 0.721 $\pm$0.002 & 0.418 $\pm$0.007 & 0.421 $\pm$0.003 & 0.384 $\pm$0.010 & 0.372 $\pm$0.010 & 0.784 $\pm$0.028 & 0.763 $\pm$0.039 & 0.846 $\pm$0.009 & 0.851 $\pm$0.005 & 0.749 $\pm$0.023 & 0.636 $\pm$0.030 & 0.257 $\pm$0.017 & 0.276 $\pm$0.008 & 0.322 $\pm$0.013 & 0.335 $\pm$0.013 & 0.656 $\pm$0.026 & 0.667 $\pm$0.020 \\
\cmidrule(lr){2-22}
 & \textbf{Random 5-shot} &  &  &  &  &  &  &  &  &  &  &  &  &  &  &  &  &  &  &  &  \\
 &   Llama-3.1-8B-Instruct & 0.650 $\pm$0.011 & 0.676 $\pm$0.008 & 0.769 $\pm$0.027 & 0.767 $\pm$0.027 & 0.467 $\pm$0.013 & 0.460 $\pm$0.016 & 0.435 $\pm$0.011 & 0.432 $\pm$0.011 & 0.727 $\pm$0.016 & 0.686 $\pm$0.019 & 0.672 $\pm$0.022 & 0.623 $\pm$0.024 & 0.587 $\pm$0.071 & 0.455 $\pm$0.076 & 0.270 $\pm$0.024 & 0.261 $\pm$0.016 & 0.260 $\pm$0.025 & 0.327 $\pm$0.051 & 0.187 $\pm$0.007 & 0.192 $\pm$0.008 \\
 &   Mistral-Nemo-Instruct-2407 & 0.761 $\pm$0.009 & 0.762 $\pm$0.009 & 0.676 $\pm$0.060 & 0.682 $\pm$0.053 & 0.425 $\pm$0.010 & 0.411 $\pm$0.011 & 0.397 $\pm$0.014 & 0.392 $\pm$0.013 & 0.611 $\pm$0.022 & 0.559 $\pm$0.021 & 0.653 $\pm$0.030 & 0.604 $\pm$0.032 & 0.783 $\pm$0.037 & 0.684 $\pm$0.052 & 0.512 $\pm$0.030 & 0.490 $\pm$0.042 & 0.309 $\pm$0.027 & 0.314 $\pm$0.031 & 0.166 $\pm$0.005 & 0.169 $\pm$0.006 \\
 &   Qwen2.5-7B-Instruct & 0.786 $\pm$0.014 & 0.788 $\pm$0.013 & 0.686 $\pm$0.017 & 0.723 $\pm$0.011 & 0.413 $\pm$0.010 & 0.417 $\pm$0.006 & 0.379 $\pm$0.014 & 0.369 $\pm$0.016 & 0.802 $\pm$0.023 & 0.786 $\pm$0.032 & 0.845 $\pm$0.004 & 0.842 $\pm$0.008 & 0.747 $\pm$0.029 & 0.634 $\pm$0.038 & 0.242 $\pm$0.015 & 0.270 $\pm$0.006 & 0.332 $\pm$0.018 & 0.339 $\pm$0.019 & \underline{\textbf{0.679 $\pm$0.029}} & \underline{\textbf{0.684 $\pm$0.023}} \\
\cmidrule(lr){2-22}
 & \textbf{Random 6-shot} &  &  &  &  &  &  &  &  &  &  &  &  &  &  &  &  &  &  &  &  \\
 &   Llama-3.1-8B-Instruct & 0.619 $\pm$0.071 & 0.657 $\pm$0.047 & 0.785 $\pm$0.025 & 0.783 $\pm$0.026 & \underline{\textbf{0.476 $\pm$0.014}} & \underline{\textbf{0.471 $\pm$0.016}} & \underline{\textbf{0.441 $\pm$0.010}} & 0.438 $\pm$0.010 & 0.730 $\pm$0.014 & 0.689 $\pm$0.017 & 0.627 $\pm$0.039 & 0.576 $\pm$0.040 & 0.567 $\pm$0.074 & 0.434 $\pm$0.076 & 0.266 $\pm$0.028 & 0.259 $\pm$0.017 & 0.265 $\pm$0.027 & 0.327 $\pm$0.059 & 0.198 $\pm$0.010 & 0.202 $\pm$0.011 \\
 &   Mistral-Nemo-Instruct-2407 & 0.754 $\pm$0.022 & 0.757 $\pm$0.020 & 0.645 $\pm$0.049 & 0.655 $\pm$0.043 & 0.419 $\pm$0.009 & 0.408 $\pm$0.012 & 0.384 $\pm$0.022 & 0.382 $\pm$0.017 & 0.600 $\pm$0.029 & 0.548 $\pm$0.029 & 0.667 $\pm$0.033 & 0.618 $\pm$0.036 & 0.780 $\pm$0.035 & 0.679 $\pm$0.050 & 0.521 $\pm$0.028 & 0.500 $\pm$0.038 & 0.302 $\pm$0.030 & 0.305 $\pm$0.032 & 0.167 $\pm$0.006 & 0.170 $\pm$0.007 \\
 &   Qwen2.5-7B-Instruct & 0.786 $\pm$0.010 & 0.786 $\pm$0.009 & 0.692 $\pm$0.013 & 0.727 $\pm$0.007 & 0.416 $\pm$0.010 & 0.421 $\pm$0.004 & 0.379 $\pm$0.009 & 0.369 $\pm$0.009 & 0.794 $\pm$0.023 & 0.775 $\pm$0.034 & 0.846 $\pm$0.007 & 0.843 $\pm$0.009 & 0.743 $\pm$0.037 & 0.630 $\pm$0.048 & 0.247 $\pm$0.026 & 0.273 $\pm$0.011 & 0.341 $\pm$0.014 & 0.348 $\pm$0.011 & 0.658 $\pm$0.023 & 0.667 $\pm$0.017 \\
\cmidrule(lr){2-22}
 & \textbf{Random 7-shot} &  &  &  &  &  &  &  &  &  &  &  &  &  &  &  &  &  &  &  &  \\
 &   Llama-3.1-8B-Instruct & 0.622 $\pm$0.097 & 0.661 $\pm$0.066 & 0.794 $\pm$0.020 & 0.792 $\pm$0.020 & 0.464 $\pm$0.015 & 0.457 $\pm$0.018 & 0.434 $\pm$0.018 & 0.432 $\pm$0.017 & 0.716 $\pm$0.013 & 0.672 $\pm$0.015 & 0.643 $\pm$0.023 & 0.592 $\pm$0.024 & 0.610 $\pm$0.078 & 0.479 $\pm$0.082 & 0.261 $\pm$0.024 & 0.257 $\pm$0.013 & 0.268 $\pm$0.017 & 0.346 $\pm$0.048 & 0.194 $\pm$0.012 & 0.199 $\pm$0.013 \\
 &   Mistral-Nemo-Instruct-2407 & 0.784 $\pm$0.014 & 0.785 $\pm$0.014 & 0.670 $\pm$0.051 & 0.676 $\pm$0.047 & 0.419 $\pm$0.010 & 0.407 $\pm$0.011 & 0.384 $\pm$0.023 & 0.383 $\pm$0.017 & 0.601 $\pm$0.029 & 0.549 $\pm$0.028 & 0.664 $\pm$0.045 & 0.616 $\pm$0.048 & 0.788 $\pm$0.035 & 0.691 $\pm$0.051 & 0.532 $\pm$0.016 & 0.516 $\pm$0.022 & 0.336 $\pm$0.017 & 0.345 $\pm$0.023 & 0.164 $\pm$0.004 & 0.168 $\pm$0.004 \\
 &   Qwen2.5-7B-Instruct & \underline{0.789 $\pm$0.009} & \underline{0.790 $\pm$0.009} & 0.683 $\pm$0.033 & 0.724 $\pm$0.018 & 0.411 $\pm$0.016 & 0.417 $\pm$0.008 & 0.383 $\pm$0.009 & 0.373 $\pm$0.010 & 0.803 $\pm$0.015 & 0.788 $\pm$0.023 & 0.844 $\pm$0.006 & 0.838 $\pm$0.010 & 0.742 $\pm$0.030 & 0.627 $\pm$0.039 & 0.214 $\pm$0.019 & 0.259 $\pm$0.008 & 0.342 $\pm$0.013 & 0.344 $\pm$0.013 & 0.659 $\pm$0.017 & 0.667 $\pm$0.013 \\
\cmidrule(lr){2-22}
 & \textbf{Random 8-shot} &  &  &  &  &  &  &  &  &  &  &  &  &  &  &  &  &  &  &  &  \\
 &   Llama-3.1-8B-Instruct & 0.590 $\pm$0.089 & 0.639 $\pm$0.053 & 0.793 $\pm$0.026 & 0.791 $\pm$0.026 & 0.466 $\pm$0.011 & 0.459 $\pm$0.014 & 0.435 $\pm$0.015 & 0.433 $\pm$0.013 & 0.703 $\pm$0.016 & 0.657 $\pm$0.018 & 0.616 $\pm$0.043 & 0.565 $\pm$0.044 & 0.531 $\pm$0.101 & 0.402 $\pm$0.098 & 0.246 $\pm$0.025 & 0.248 $\pm$0.014 & 0.258 $\pm$0.022 & 0.321 $\pm$0.050 & 0.225 $\pm$0.022 & 0.230 $\pm$0.026 \\
 &   Mistral-Nemo-Instruct-2407 & 0.770 $\pm$0.014 & 0.771 $\pm$0.013 & 0.653 $\pm$0.049 & 0.661 $\pm$0.043 & 0.414 $\pm$0.011 & 0.400 $\pm$0.012 & 0.392 $\pm$0.017 & 0.389 $\pm$0.014 & 0.608 $\pm$0.030 & 0.556 $\pm$0.030 & 0.661 $\pm$0.028 & 0.612 $\pm$0.030 & 0.787 $\pm$0.041 & 0.690 $\pm$0.059 & 0.522 $\pm$0.014 & 0.500 $\pm$0.019 & 0.342 $\pm$0.022 & 0.355 $\pm$0.025 & 0.177 $\pm$0.006 & 0.182 $\pm$0.006 \\
 &   Qwen2.5-7B-Instruct & 0.780 $\pm$0.009 & 0.780 $\pm$0.009 & 0.685 $\pm$0.034 & 0.724 $\pm$0.022 & 0.412 $\pm$0.013 & 0.416 $\pm$0.007 & 0.375 $\pm$0.004 & 0.364 $\pm$0.004 & 0.795 $\pm$0.014 & 0.775 $\pm$0.018 & 0.844 $\pm$0.005 & 0.837 $\pm$0.009 & 0.748 $\pm$0.030 & 0.635 $\pm$0.040 & 0.208 $\pm$0.031 & 0.259 $\pm$0.013 & \underline{\textbf{0.364 $\pm$0.032}} & 0.366 $\pm$0.032 & 0.646 $\pm$0.029 & 0.656 $\pm$0.024 \\
\cmidrule(lr){2-22}
 & \textbf{Random 9-shot} &  &  &  &  &  &  &  &  &  &  &  &  &  &  &  &  &  &  &  &  \\
 &   Llama-3.1-8B-Instruct & 0.631 $\pm$0.060 & 0.665 $\pm$0.041 & \underline{\textbf{0.811 $\pm$0.013}} & \underline{\textbf{0.809 $\pm$0.013}} & 0.468 $\pm$0.017 & 0.462 $\pm$0.022 & 0.440 $\pm$0.021 & \underline{\textbf{0.439 $\pm$0.021}} & 0.711 $\pm$0.022 & 0.667 $\pm$0.025 & 0.645 $\pm$0.034 & 0.595 $\pm$0.036 & 0.578 $\pm$0.092 & 0.447 $\pm$0.097 & 0.235 $\pm$0.027 & 0.242 $\pm$0.014 & 0.290 $\pm$0.014 & \underline{0.384 $\pm$0.019} & 0.219 $\pm$0.016 & 0.229 $\pm$0.024 \\
 &   Mistral-Nemo-Instruct-2407 & 0.780 $\pm$0.030 & 0.781 $\pm$0.028 & 0.690 $\pm$0.052 & 0.694 $\pm$0.046 & 0.413 $\pm$0.010 & 0.400 $\pm$0.011 & 0.388 $\pm$0.017 & 0.385 $\pm$0.014 & 0.607 $\pm$0.025 & 0.555 $\pm$0.024 & 0.671 $\pm$0.037 & 0.623 $\pm$0.040 & 0.796 $\pm$0.047 & 0.704 $\pm$0.067 & 0.523 $\pm$0.009 & 0.501 $\pm$0.013 & 0.326 $\pm$0.029 & 0.341 $\pm$0.026 & 0.170 $\pm$0.008 & 0.174 $\pm$0.008 \\
 &   Qwen2.5-7B-Instruct & 0.780 $\pm$0.014 & 0.781 $\pm$0.013 & 0.696 $\pm$0.021 & 0.732 $\pm$0.014 & 0.417 $\pm$0.009 & 0.416 $\pm$0.009 & 0.384 $\pm$0.008 & 0.373 $\pm$0.008 & 0.794 $\pm$0.011 & 0.773 $\pm$0.016 & 0.846 $\pm$0.008 & 0.840 $\pm$0.014 & 0.743 $\pm$0.040 & 0.630 $\pm$0.051 & 0.208 $\pm$0.016 & 0.256 $\pm$0.008 & 0.359 $\pm$0.026 & 0.362 $\pm$0.027 & 0.659 $\pm$0.031 & 0.668 $\pm$0.025 \\
\cmidrule(lr){2-22}
 & \textbf{Random 10-shot} &  &  &  &  &  &  &  &  &  &  &  &  &  &  &  &  &  &  &  &  \\
 &   Llama-3.1-8B-Instruct & 0.587 $\pm$0.045 & 0.635 $\pm$0.029 & 0.798 $\pm$0.019 & 0.796 $\pm$0.019 & 0.462 $\pm$0.013 & 0.457 $\pm$0.017 & 0.432 $\pm$0.019 & 0.431 $\pm$0.019 & 0.694 $\pm$0.019 & 0.647 $\pm$0.021 & 0.612 $\pm$0.036 & 0.561 $\pm$0.037 & 0.582 $\pm$0.100 & 0.453 $\pm$0.103 & 0.243 $\pm$0.026 & 0.245 $\pm$0.012 & 0.285 $\pm$0.026 & 0.367 $\pm$0.048 & 0.254 $\pm$0.012 & 0.273 $\pm$0.019 \\
 &   Mistral-Nemo-Instruct-2407 & 0.771 $\pm$0.016 & 0.772 $\pm$0.015 & 0.691 $\pm$0.053 & 0.695 $\pm$0.049 & 0.414 $\pm$0.012 & 0.401 $\pm$0.012 & 0.391 $\pm$0.019 & 0.388 $\pm$0.016 & 0.615 $\pm$0.028 & 0.563 $\pm$0.028 & 0.679 $\pm$0.027 & 0.631 $\pm$0.030 & 0.790 $\pm$0.060 & 0.697 $\pm$0.084 & 0.500 $\pm$0.007 & 0.470 $\pm$0.009 & 0.352 $\pm$0.045 & 0.364 $\pm$0.050 & 0.183 $\pm$0.005 & 0.190 $\pm$0.008 \\
 &   Qwen2.5-7B-Instruct & 0.783 $\pm$0.010 & 0.783 $\pm$0.010 & 0.704 $\pm$0.024 & 0.736 $\pm$0.016 & 0.425 $\pm$0.008 & 0.421 $\pm$0.007 & 0.384 $\pm$0.014 & 0.373 $\pm$0.014 & 0.786 $\pm$0.011 & 0.760 $\pm$0.016 & \underline{0.847 $\pm$0.006} & 0.841 $\pm$0.010 & 0.750 $\pm$0.048 & 0.640 $\pm$0.063 & 0.208 $\pm$0.018 & 0.256 $\pm$0.008 & 0.361 $\pm$0.014 & 0.362 $\pm$0.013 & 0.236 $\pm$0.020 & 0.237 $\pm$0.021 \\
\cmidrule(lr){2-22}
 & \textbf{DPP 2-shot} &  &  &  &  &  &  &  &  &  &  &  &  &  &  &  &  &  &  &  &  \\
 &   Llama-3.1-8B-Instruct & 0.685 $\pm$0.063 & 0.704 $\pm$0.048 & 0.787 $\pm$0.010 & 0.784 $\pm$0.010 & - & - & - & - & 0.749 $\pm$0.010 & 0.712 $\pm$0.012 & 0.695 $\pm$0.053 & 0.650 $\pm$0.057 & 0.409 $\pm$0.054 & 0.292 $\pm$0.042 & 0.274 $\pm$0.025 & 0.261 $\pm$0.017 & 0.276 $\pm$0.032 & 0.309 $\pm$0.035 & 0.209 $\pm$0.024 & 0.226 $\pm$0.041 \\
 &   Mistral-Nemo-Instruct-2407 & 0.625 $\pm$0.039 & 0.660 $\pm$0.027 & 0.626 $\pm$0.046 & 0.640 $\pm$0.038 & - & - & - & - & 0.664 $\pm$0.030 & 0.614 $\pm$0.032 & 0.608 $\pm$0.093 & 0.562 $\pm$0.090 & 0.785 $\pm$0.022 & 0.686 $\pm$0.031 & 0.524 $\pm$0.056 & 0.511 $\pm$0.074 & 0.264 $\pm$0.006 & 0.276 $\pm$0.017 & 0.155 $\pm$0.009 & 0.158 $\pm$0.011 \\
 &   Qwen2.5-7B-Instruct & 0.782 $\pm$0.010 & 0.784 $\pm$0.009 & 0.633 $\pm$0.019 & 0.694 $\pm$0.010 & - & - & - & - & 0.818 $\pm$0.010 & 0.814 $\pm$0.017 & 0.836 $\pm$0.007 & 0.848 $\pm$0.010 & 0.746 $\pm$0.019 & 0.633 $\pm$0.025 & 0.205 $\pm$0.039 & 0.258 $\pm$0.012 & 0.278 $\pm$0.017 & 0.298 $\pm$0.011 & 0.205 $\pm$0.020 & 0.213 $\pm$0.025 \\
\cmidrule(lr){2-22}
 & \textbf{DPP 3-shot} &  &  &  &  &  &  &  &  &  &  &  &  &  &  &  &  &  &  &  &  \\
 &   Llama-3.1-8B-Instruct & - & - & - & - & 0.460 $\pm$0.011 & 0.453 $\pm$0.013 & 0.422 $\pm$0.010 & 0.416 $\pm$0.011 & - & - & - & - & - & - & - & - & - & - & - & - \\
 &   Mistral-Nemo-Instruct-2407 & - & - & - & - & 0.422 $\pm$0.008 & 0.417 $\pm$0.010 & 0.394 $\pm$0.008 & 0.386 $\pm$0.006 & - & - & - & - & - & - & - & - & - & - & - & - \\
 &   Qwen2.5-7B-Instruct & - & - & - & - & 0.401 $\pm$0.003 & 0.407 $\pm$0.012 & 0.359 $\pm$0.012 & 0.350 $\pm$0.010 & - & - & - & - & - & - & - & - & - & - & - & - \\
\cmidrule(lr){2-22}
 & \textbf{DPP 4-shot} &  &  &  &  &  &  &  &  &  &  &  &  &  &  &  &  &  &  &  &  \\
 &   Llama-3.1-8B-Instruct & 0.708 $\pm$0.056 & 0.721 $\pm$0.043 & 0.771 $\pm$0.021 & 0.768 $\pm$0.021 & - & - & - & - & 0.730 $\pm$0.018 & 0.690 $\pm$0.022 & 0.669 $\pm$0.022 & 0.619 $\pm$0.024 & 0.415 $\pm$0.079 & 0.298 $\pm$0.061 & 0.262 $\pm$0.032 & 0.253 $\pm$0.021 & 0.246 $\pm$0.019 & 0.264 $\pm$0.025 & 0.219 $\pm$0.044 & 0.238 $\pm$0.075 \\
 &   Mistral-Nemo-Instruct-2407 & 0.713 $\pm$0.021 & 0.724 $\pm$0.015 & 0.563 $\pm$0.031 & 0.590 $\pm$0.022 & - & - & - & - & 0.649 $\pm$0.013 & 0.598 $\pm$0.014 & 0.687 $\pm$0.012 & 0.639 $\pm$0.013 & 0.781 $\pm$0.041 & 0.681 $\pm$0.057 & 0.516 $\pm$0.018 & 0.496 $\pm$0.026 & 0.274 $\pm$0.019 & 0.279 $\pm$0.019 & 0.160 $\pm$0.008 & 0.164 $\pm$0.008 \\
 &   Qwen2.5-7B-Instruct & 0.781 $\pm$0.005 & 0.782 $\pm$0.006 & 0.640 $\pm$0.013 & 0.695 $\pm$0.008 & - & - & - & - & 0.806 $\pm$0.013 & 0.793 $\pm$0.019 & 0.840 $\pm$0.014 & 0.841 $\pm$0.023 & 0.749 $\pm$0.025 & 0.636 $\pm$0.033 & 0.220 $\pm$0.029 & 0.259 $\pm$0.010 & 0.330 $\pm$0.024 & 0.343 $\pm$0.017 & 0.224 $\pm$0.014 & 0.231 $\pm$0.017 \\
\cmidrule(lr){2-22}
 & \textbf{DPP 6-shot} &  &  &  &  &  &  &  &  &  &  &  &  &  &  &  &  &  &  &  &  \\
 &   Llama-3.1-8B-Instruct & 0.665 $\pm$0.047 & 0.686 $\pm$0.032 & 0.793 $\pm$0.023 & 0.791 $\pm$0.023 & 0.461 $\pm$0.022 & 0.455 $\pm$0.021 & 0.433 $\pm$0.013 & 0.429 $\pm$0.017 & 0.728 $\pm$0.029 & 0.688 $\pm$0.034 & 0.658 $\pm$0.035 & 0.608 $\pm$0.037 & 0.466 $\pm$0.041 & 0.337 $\pm$0.035 & 0.230 $\pm$0.022 & 0.235 $\pm$0.013 & 0.277 $\pm$0.031 & 0.291 $\pm$0.024 & 0.219 $\pm$0.017 & 0.226 $\pm$0.025 \\
 &   Mistral-Nemo-Instruct-2407 & 0.682 $\pm$0.026 & 0.698 $\pm$0.021 & 0.530 $\pm$0.041 & 0.566 $\pm$0.028 & 0.410 $\pm$0.010 & 0.410 $\pm$0.015 & 0.390 $\pm$0.012 & 0.381 $\pm$0.011 & 0.630 $\pm$0.013 & 0.578 $\pm$0.015 & 0.689 $\pm$0.030 & 0.643 $\pm$0.032 & 0.747 $\pm$0.052 & 0.636 $\pm$0.070 & 0.510 $\pm$0.039 & 0.490 $\pm$0.056 & 0.302 $\pm$0.029 & 0.304 $\pm$0.028 & 0.162 $\pm$0.007 & 0.170 $\pm$0.007 \\
 &   Qwen2.5-7B-Instruct & 0.779 $\pm$0.010 & 0.780 $\pm$0.010 & 0.632 $\pm$0.013 & 0.691 $\pm$0.008 & 0.406 $\pm$0.006 & 0.413 $\pm$0.009 & 0.354 $\pm$0.009 & 0.345 $\pm$0.008 & 0.795 $\pm$0.022 & 0.777 $\pm$0.033 & 0.843 $\pm$0.011 & 0.840 $\pm$0.017 & 0.724 $\pm$0.024 & 0.604 $\pm$0.030 & 0.190 $\pm$0.028 & 0.249 $\pm$0.009 & 0.338 $\pm$0.028 & 0.352 $\pm$0.024 & 0.235 $\pm$0.003 & 0.242 $\pm$0.006 \\
\cmidrule(lr){2-22}
 & \textbf{DPP 8-shot} &  &  &  &  &  &  &  &  &  &  &  &  &  &  &  &  &  &  &  &  \\
 &   Llama-3.1-8B-Instruct & 0.692 $\pm$0.013 & 0.703 $\pm$0.011 & 0.787 $\pm$0.011 & 0.786 $\pm$0.012 & - & - & - & - & 0.707 $\pm$0.033 & 0.663 $\pm$0.036 & 0.619 $\pm$0.034 & 0.568 $\pm$0.035 & 0.498 $\pm$0.057 & 0.367 $\pm$0.052 & 0.216 $\pm$0.027 & 0.227 $\pm$0.012 & 0.292 $\pm$0.016 & 0.303 $\pm$0.020 & 0.257 $\pm$0.043 & 0.287 $\pm$0.079 \\
 &   Mistral-Nemo-Instruct-2407 & 0.702 $\pm$0.029 & 0.710 $\pm$0.025 & 0.536 $\pm$0.047 & 0.571 $\pm$0.033 & - & - & - & - & 0.626 $\pm$0.012 & 0.575 $\pm$0.013 & 0.721 $\pm$0.023 & 0.678 $\pm$0.026 & 0.823 $\pm$0.033 & 0.742 $\pm$0.050 & 0.531 $\pm$0.041 & 0.521 $\pm$0.063 & 0.345 $\pm$0.039 & 0.352 $\pm$0.041 & 0.177 $\pm$0.007 & 0.187 $\pm$0.006 \\
 &   Qwen2.5-7B-Instruct & 0.767 $\pm$0.012 & 0.767 $\pm$0.012 & 0.627 $\pm$0.012 & 0.688 $\pm$0.007 & - & - & - & - & 0.793 $\pm$0.014 & 0.772 $\pm$0.021 & 0.847 $\pm$0.014 & 0.841 $\pm$0.020 & 0.715 $\pm$0.035 & 0.593 $\pm$0.045 & 0.170 $\pm$0.025 & 0.243 $\pm$0.008 & 0.349 $\pm$0.019 & 0.362 $\pm$0.028 & 0.234 $\pm$0.015 & 0.236 $\pm$0.016 \\
\cmidrule(lr){2-22}
 & \textbf{DPP 9-shot} &  &  &  &  &  &  &  &  &  &  &  &  &  &  &  &  &  &  &  &  \\
 &   Llama-3.1-8B-Instruct & - & - & - & - & 0.459 $\pm$0.017 & 0.454 $\pm$0.019 & 0.433 $\pm$0.014 & 0.428 $\pm$0.014 & - & - & - & - & - & - & - & - & - & - & - & - \\
 &   Mistral-Nemo-Instruct-2407 & - & - & - & - & 0.405 $\pm$0.006 & 0.407 $\pm$0.009 & 0.383 $\pm$0.011 & 0.376 $\pm$0.011 & - & - & - & - & - & - & - & - & - & - & - & - \\
 &   Qwen2.5-7B-Instruct & - & - & - & - & 0.412 $\pm$0.007 & 0.416 $\pm$0.009 & 0.362 $\pm$0.013 & 0.353 $\pm$0.011 & - & - & - & - & - & - & - & - & - & - & - & - \\
\cmidrule(lr){2-22}
 & \textbf{DPP 10-shot} &  &  &  &  &  &  &  &  &  &  &  &  &  &  &  &  &  &  &  &  \\
 &   Llama-3.1-8B-Instruct & 0.696 $\pm$0.017 & 0.708 $\pm$0.013 & 0.801 $\pm$0.020 & 0.800 $\pm$0.021 & - & - & - & - & 0.681 $\pm$0.042 & 0.634 $\pm$0.044 & 0.653 $\pm$0.036 & 0.603 $\pm$0.037 & 0.523 $\pm$0.060 & 0.390 $\pm$0.059 & 0.202 $\pm$0.016 & 0.220 $\pm$0.005 & 0.269 $\pm$0.029 & 0.280 $\pm$0.034 & 0.247 $\pm$0.028 & 0.267 $\pm$0.042 \\
 &   Mistral-Nemo-Instruct-2407 & 0.687 $\pm$0.028 & 0.694 $\pm$0.020 & 0.542 $\pm$0.064 & 0.576 $\pm$0.046 & - & - & - & - & 0.643 $\pm$0.017 & 0.592 $\pm$0.018 & 0.753 $\pm$0.022 & 0.715 $\pm$0.026 & \underline{0.851 $\pm$0.041} & \underline{0.788 $\pm$0.066} & 0.519 $\pm$0.036 & 0.502 $\pm$0.052 & 0.332 $\pm$0.046 & 0.346 $\pm$0.058 & 0.178 $\pm$0.008 & 0.186 $\pm$0.008 \\
 &   Qwen2.5-7B-Instruct & 0.759 $\pm$0.011 & 0.760 $\pm$0.011 & 0.627 $\pm$0.021 & 0.688 $\pm$0.012 & - & - & - & - & 0.794 $\pm$0.017 & 0.774 $\pm$0.024 & 0.846 $\pm$0.015 & 0.841 $\pm$0.022 & 0.724 $\pm$0.036 & 0.604 $\pm$0.044 & 0.159 $\pm$0.023 & 0.239 $\pm$0.005 & 0.356 $\pm$0.041 & 0.368 $\pm$0.050 & 0.229 $\pm$0.018 & 0.230 $\pm$0.019 \\
\midrule
\midrule
\multirow{4}{*}{\parbox[c][4\baselineskip][c]{1cm}{\centering\rotatebox{90}{\textbf{CoT}}}} & \textbf{CoT} &  &  &  &  &  &  &  &  &  &  &  &  &  &  &  &  &  &  &  &  \\
 &   Llama-3.1-8B-Instruct & \underline{0.792} & \underline{0.795} & \underline{0.757} & \underline{0.764} & \underline{0.465} & 0.459 & \underline{0.416} & \underline{0.405} & 0.557 & 0.502 & 0.503 & 0.454 & 0.226 & 0.160 & 0.377 & 0.340 & 0.280 & 0.358 & 0.161 & 0.161 \\
 &   Mistral-Nemo-Instruct-2407 & 0.713 & 0.728 & 0.732 & 0.729 & 0.464 & \underline{0.465} & 0.401 & 0.386 & 0.252 & 0.275 & 0.584 & 0.530 & 0.528 & 0.394 & \underline{\textbf{0.623}} & \underline{\textbf{0.680}} & \underline{0.315} & \underline{0.397} & \underline{0.167} & \underline{0.168} \\
 &   Qwen2.5-7B-Instruct & 0.711 & 0.725 & 0.714 & 0.729 & 0.422 & 0.428 & 0.355 & 0.349 & \underline{0.575} & \underline{0.522} & \underline{0.804} & \underline{0.783} & \underline{0.646} & \underline{0.511} & 0.491 & 0.456 & 0.277 & 0.340 & 0.147 & 0.149 \\
\bottomrule
\end{tabular}}%

\caption{Comparison of F1 and accuracy across all models, configurations, and datasets. Each cell shows the mean F1 and accuracy ($\pm$ one standard deviation for Few-Shot only, since Zero-Shot and CoT use greedy decoding and have no standard deviation). \textbf{Boldface} = best overall; \underline{underlined} = best within each section.}
    \label{tab:performance_comparison_portrait_bold}
  \end{minipage}%
}
\end{table}

\begin{table*}[h!]
\resizebox{\linewidth}{!}{%
\begin{tabular}{ccccccccccccccccccccc}
 \hline
 Dataset & {Reference} & {Model} & \multicolumn{2}{c}{Performance} & {Our Best Model}& {Configuration} &\multicolumn{2}{c}{Performance} \\
 & & & Acc & F1 & & & Acc & F1 \\
 
 \hline
 VIStA-H & 
 \cite{lyu_computational_2023} & 
 BERT-base& 
 0.84 & 0.78 &
 RoBERTa-large & FT &
 \textbf{.852} & \textbf{.850} & 
 \\

 Fake News Net & 
 \cite{jinFineGrainedReasoningFake2022} & Graph-based Reasoning & 
0.870 & 0.892 & Llama3.1-8b & FT & .945 & .945 \\

 Spanish Fake News Corpus & \cite{gomez2021overview} & BERT& 0.766 & N/A & Modern-BERT-large & FT & \textbf{.863} & \textbf{.863}\\

 Fake.br & \cite{fakebr} & SVM & 0.89 & 0.89 & Llama3-8b-Instruct & FT & \textbf{.979} & \textbf{.979} \\

 \hline
\end{tabular}%
}
\caption{SOTA results.}\label{tab:SOTAvsOURS}
\end{table*}

\end{document}